\documentclass[journal]{IEEEtran}
\usepackage{graphicx}
\usepackage{amsmath}
\usepackage{amssymb}
\usepackage{booktabs}
\usepackage{multirow}
\usepackage{multicol}
 \usepackage{array}
 \usepackage{float}
\usepackage{mdframed}
\usepackage[skip=1pt]{caption}
\usepackage{lipsum}
\newcommand{\STAB}[1]{\begin{tabular}{@{}c@{}}#1\end{tabular}}

\usepackage[pagebackref,breaklinks,colorlinks]{hyperref}

\begin{document}
\title{Fast Yet Effective Machine Unlearning}

\makeatletter
\newcommand{\printfnsymbol}[1]{%
  \textsuperscript{\@fnsymbol{#1}}%
}
\makeatother

\author{Ayush~K~Tarun\textsuperscript{$\dagger$}, Vikram~S~Chundawat\textsuperscript{$\dagger$}, Murari~Mandal{$\ddagger$}, Mohan~Kankanhalli
\thanks{\textsuperscript{$\dagger$}Equal Contribution. The work is part of the authors' internship at the School of Computing, National University of Singapore.} 
\thanks{\textsuperscript{$\ddagger$}\textit{Corresponding Author}. Work performed while at the School of Computing, National University of Singapore.}
\thanks{Ayush~K~Tarun, Vikram~S~Chundawat are with Mavvex Labs, India, Faridabad 121001 (ayushtarun210@gmail.com; vikram2000b@gmail.com), Murari~Mandal is with the School of Computer Engineering, Kalinga Institute of Industrial Technology (KIIT) Bhubaneswar, India 751024 (Email: murari.mandalfcs@kiit.ac.in), Mohan~Kankanhalli is with the School of Computing, National University of Singapore (NUS), Singapore 117417 (Email: mohan@comp.nus.edu.sg)}
}
\maketitle
\begin{abstract}
Unlearning the data observed during the training of a machine learning (ML) model is an important task that can play a pivotal role in fortifying the privacy and security of ML-based applications. This paper raises the following questions: (i) can we unlearn a single or multiple class(es) of data from a ML model without looking at the full training data even once? (ii) can we make the process of unlearning fast and scalable to large datasets, and generalize it to different deep networks? We introduce a novel machine unlearning framework with error-maximizing noise generation and impair-repair based weight manipulation that offers an efficient solution to the above questions. An error-maximizing noise matrix is learned for the class to be unlearned using the original model. The noise matrix is used to manipulate the model weights to unlearn the targeted class of data. We introduce impair and repair steps for a controlled manipulation of the network weights. In the impair step, the noise matrix along with a very high learning rate is used to induce sharp unlearning in the model. Thereafter, the repair step is used to regain the overall performance. With very few update steps, we show excellent unlearning while substantially retaining the overall model accuracy. Unlearning multiple classes requires a similar number of update steps as for a single class, making our approach scalable to large problems. Our method is quite efficient in comparison to the existing methods, works for multi-class unlearning, does not put any constraints on the original optimization mechanism or network design, and works well in both small and large-scale vision tasks. This work is an important step towards fast and easy implementation of unlearning in deep networks. Source code:~\url{https://github.com/vikram2000b/Fast-Machine-Unlearning}
\end{abstract}
\IEEEpeerreviewmaketitle
\section{Introduction}
\label{sec:intro}
Consider a scenario where it is desired that the information pertaining to the data belonging to a single class or multiple classes be removed from the already trained machine learning (ML) model. For example, a company is requested to remove the face image data for a user (or a set of users) from the already trained face recognition model. In addition, there is a constraint such that the company no longer has access to those (requested to be removed) facial images. 
How do we solve such a problem? With the increase in privacy awareness among the general populace and the cognizance of the negative impacts of sharing one's data with ML-based applications, such type of demands could be raised frequently in near future. Privacy regulations~\cite{voigt2017eu,goldman2020introduction} are increasingly likely to include such provisions in future to give the control of personal privacy to the individuals. For example, the California Consumer Privacy Act (CCPA)~\cite{goldman2020introduction} allows companies to collect user data by default. However, the user has the \textit{right to delete} her personal data and \textit{right to opt-out} of the sale of her personal information. In case a company has already used the data collected from the users (in our example, \textit{face data}) to train its ML model, then the model needs to be manipulated suitably to reflect the data deletion request. The naive way is to redo the model training from scratch for every such request. This would result in significant cost of time and resources to the company. How can this process be made more efficient? What are the challenges? How do we know that the model has actually unlearned those class/classes of data? How to ensure minimal effect on the overall accuracy of the model? 
These are some of the questions that have been asked and possible solutions have been explored in recent times~\cite{golatkar2021mixed,bourtoule2021machine,sommer2022athena,garg2020formalizing,brophy2021machine,nguyen2020variational,golatkar2020eternal,golatkar2020forgetting,ginart2019making,guo2020certified,baumhauer2022machine,izzo2021approximate,neel2021descent,wu2020deltagrad,chunda2023badteach,tarun2022deep,chundawat2022zero}.\par

The unlearning (also called selective forgetting, data deletion, or scrubbing) solutions presented in the literature are focused on simple learning algorithms such a linear/logistic regression~\cite{mahadevan2021certifiable}, random forests~\cite{brophy2021machine}, and k-means clustering~\cite{ginart2019making}. Initial work on forgetting in convolutional networks is presented in~\cite{golatkar2020eternal,golatkar2020forgetting}. However, these methods are shown to be effective only on small scale problems and are computationally expensive. Efficient unlearning in deep networks such as CNN and Vision Transformers still remain an open problem. In particular, efficiently unlearning of multiple classes is yet to be explored. This is due to several complexities that arise while working with deep learning models. For example, the non-convex loss space~\cite{choromanska2015loss} of CNNs makes it difficult to assess the effect of a data sample on the optimization trajectory and the final network weight combination. Furthermore, several optimal set of weights may exist for the same network, making it difficult to confidently evaluate the degree of unlearning. Comparing the updated model weights after unlearning with a model trained without the forget classes might not reveal helpful information on the quality of unlearning. Forgetting a cohort of data or an entire class of data while preserving the accuracy of the model is a non-trivial problem as has been shown in the existing works~\cite{golatkar2021mixed,golatkar2020eternal}. Moreover, efficiently manipulating the network weights without using the unlearning data still remains an unsolved problem. Other challenges are to unlearn multiple classes of data, perform unlearning for large-scale problems, and generalize the solution to different types of deep networks.\par
Estimating the effect of a data sample or a class of data samples on the deep model parameters is a challenging problem~\cite{koh2017understanding,koh2019accuracy}. Therefore, several unlearning research efforts have been focused on the simpler convex learning problems (i.e., linear/logistic regression) that offer better theoretical analysis. Researchers have~\cite{koh2017understanding} used influence functions to study the behaviour of black-box models such as CNNs through the lens of training samples. It is observed that data points with high training loss are more influential for the model parameters. The adversarial versions~\cite{FGSM} of the training images are generated by maximizing the loss on these images. It is further shown that the influence functions are also useful for studying the effect of a group of data points~\cite{koh2019accuracy}. Recently, Huang et al.~\cite{huang2021unlearnable} proposed to learn an error-minimizing noise to make training examples unlearnable for deep learning models. The idea is to add such noise to the image samples that fools the model in believing nothing is to be learned from those samples. If used in training, such images have no effect on the model.

Unlearning requires the model to forget specific class(es) of data but remember the rest of the data. For the class(es) to be forgotten, if the model can be updated by observing patterns that are somehow \emph{opposites} of the patterns learned at the time of original training, then the updated model weights might reflect the desired unlearning. And hopefully it preserves the remaining classes information. We know that the original model is trained by minimizing the loss for all the classes. So intuitively, maximizing a noise with respect to the model loss only for the unlearning class will help us learn such patterns that help forgetting. It can also be  viewed as learning anti-samples for a given class and use these anti-samples to damage the previously learned information. In this paper, we propose a framework for unlearning in a \textit{zero-glance} privacy setting, i.e. the model can not see the unlearning class of data. We learn an error-maximizing noise matrix consisting of highly influential points corresponding to the unlearning class. After that, we train the model using the noise matrix to update the network weights. We introduce Unlearning by Selective Impair and Repair (UNSIR), a \textit{single-pass} method to unlearn single/multiple classes of data in a deep model without requiring access to the data samples of the requested set of unlearning classes. Our method can be directly applied on the already trained deep model to make it forget the information about the requested class of data - while at the same time retaining very close to the original accuracy of the model on the remaining tasks. In fact our method performs exceedingly well in both unlearning the requested classes and retaining the accuracy on the remaining classes. To the best of our knowledge, this is the first method to achieve efficient multi-class unlearning in deep networks not only for small-scale problems (10 classes) but also for large-scale vision problems (100 classes). Our method works with the stringent \textit{zero-glance}
setting where \textit{data samples of the requested unlearning class is either not available or can not be used}. This makes our solution unique and practical for real-world application. An important and realistic use-case of unlearning is face recognition. We show that our method can effectively make a trained model forget a single as well as multiple faces in a highly efficient manner, without glancing at the samples of the unlearning faces.\par

To summarize, our key contributions are:
\begin{enumerate}
    \item We introduce the problem of unlearning in a \textit{zero-glance} setting which is a stricter formalization compared to the existing settings and offers a prospect for higher-level of privacy guarantees.
    
    \item We learn an error-maximizing noise for the respective unlearning classes. \textit{UNSIR} is proposed to perform single-pass impair and single-pass repair by using a very high learning rate. The impair step makes the network forget the unlearning data classes. The repair step stabilizes the network weights to better remember the remaining tasks. The combination of both the steps allows it to obtain excellent unlearning and retain accuracy. 

    \item We show that along with a better privacy setting and offering multi-class unlearning, our method is also highly efficient. The multi-class unlearning is performed in a single impair-repair pass instead of sequentially unlearning individual classes.
    
    \item The proposed method works on large-scale vision datasets with strong performance on different types of deep networks such as convolutional networks and Vision Transformers. Our method does not require any prior information related to process of original model training and it is easily applicable to a wide class of deep networks. Specifically, we show excellent unlearning results on face recognition. To the best of our knowledge, it is the first machine unlearning method to demonstrate all the above characteristics together.
\end{enumerate}

\section{Related Work}
\label{sec:related}
\subsection{Machine Unlearning} 
Machine unlearning was formulated as a data forgetting algorithm in statistical query learning~\cite{cao2015towards}. Brophy and Lowd~\cite{brophy2021machine} introduced a variant of random forests that supports data forgetting with minimal retraining. Data deletion in k-means clustering has been studied in~\cite{ginart2019making,mirzasoleiman2017deletion}. Guo et al.~\cite{guo2020certified} give a certified information removal framework based on Newton's update removal mechanism for convex learning problems. The data removal is certified using a variation of the differential privacy condition~\cite{abadi2016deep,dwork2014algorithmic}. Izzo et al.~\cite{izzo2021approximate} presents a projective residual update method to delete data points from linear models. A method to hide the class information from the output logits is presented in~\cite{baumhauer2020machine}. This however, does not remove the information present in the network weights. Unlearning in a Bayesian setting using variational inference is explored for regression and Gaussian processes in~\cite{nguyen2020variational}. Neel et al.~\cite{neel2021descent} study the results of gradient descent based approach to unlearning in convex models. All these methods are designed for convex problems, whereas we aim to present an unlearning solution for deep learning models.\par 

Some methods adopt strategic grouping of data in the training procedure and thus enable smooth unlearning by limiting the influence of data points on model learning~\cite{bourtoule2021machine,wu2020deltagrad}. This approach results in high storage cost as it mandates storing multiple snapshots of the network and gradients to ensure good unlearning performance. These approaches are independent of the types of learning algorithms and rely on the efficient division of training data. They also need to retrain a subset of the models, while we aim to create a highly efficient unlearning algorithm without any memory overhead. Gupta et al.~\cite{gupta2021adaptive} proposed an algorithm to handle a sequence of adaptive deletion requests in this setting.  

\begin{figure*}[]
    \centering
    \includegraphics[width=0.9\linewidth]{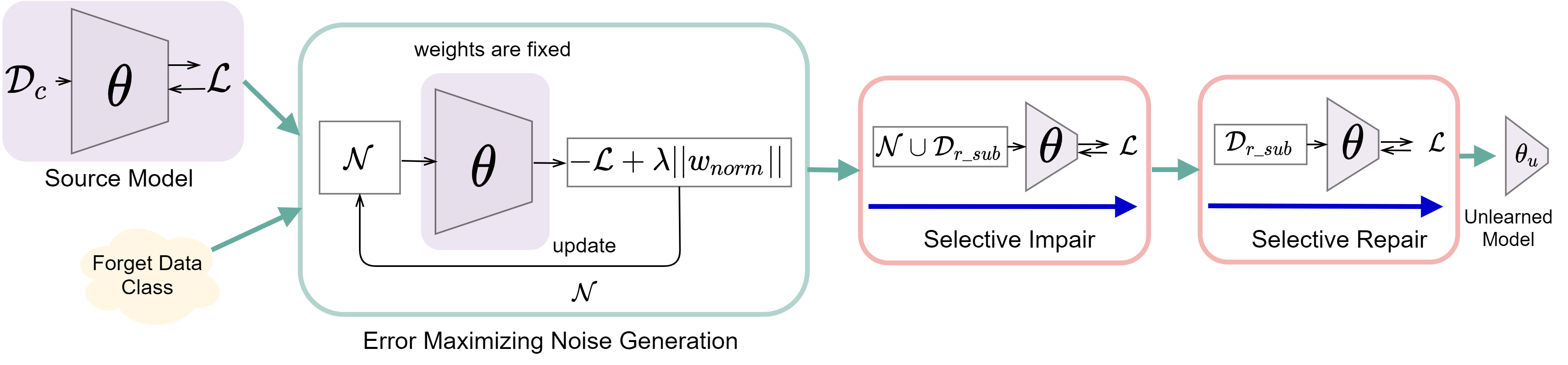}
    \caption{
    The proposed unlearning framework. We use the pretrained model to learn the error-maximizing noise matrix for the unlearning class. The generated noise $\mathcal{N}$ is then used along with a subset of the retain data $\mathcal{D}_{r\_sub}$ to update the model with one epoch (impair). Next, we apply a healing step by further updating the network with only the retain data $\mathcal{D}_{r\_sub}$ (repair). The repair step helps in regaining the overall model performance while unlearning the requested class/classes of data.}
    \label{fig:framework}
\end{figure*}

\subsection{Unlearning in Deep Neural Networks}
Forgetting in deep neural networks is challenging due to their highly non-convex loss functions. Although the term \textit{forgetting} is used quite often in continual learning literature~\cite{prabhu2020gdumb}, where a model rapidly loses accuracy on the previous task when fine-tuned for a new task. This however does not address the information remaining in the network weights. Throughout this paper we use the term \textit{unlearning} and \textit{forgetting} interchangeably, both denoting that the information of data in the network weights are also removed. Golatkar et al.~\cite{golatkar2020eternal} proposed an information theoretic method to scrub the information from intermediate layers of deep networks trained with stochastic gradient descent (SGD). They also give an upper-bound on the amount of remaining information in the network~\cite{achille2019information} after forgetting by exploiting the stability of SGD. This work is extended~\cite{golatkar2020forgetting} by including an update mechanism for the final activations of the model. They present a neural tangent kernel (NTK) based approximation of the training process and use it to estimate the updated network weights after forgetting. However, both the approximation accuracy and computational costs degrade for larger datasets. The computational cost even in a small dataset is quite high as the cost is quadratic in the number of samples. In~\cite{golatkar2021mixed}, the authors directly train a linearized network and use it for forgetting. They train two separate networks: the core model, and a mixed-linear model. The mixed-linear model requires Jacobian-vector product (JVP) computation and few other fine-tuning. This framework was shown to be scalable for several standard vision datasets. However, they present such a network only for ResNet50 which requires a lot of fine-tuning to obtain the results. Also designing a mixed-linear network for every deep architecture is an inefficient approach. Some researchers have studied the unintended privacy risks resulting from the existing unlearning methods~\cite{chen2021machine,marchant2021hard}.~\cite{thudi2021necessity} show the difficulty of formally proving the absence of certain data points in the model. They suggest that the current unlearning methods are well-defined only at the algorithmic level. Forgetting in federated learning~\cite{wu2022federated} and recommendation systems~\cite{chen2022recommendation} are also explored. Several other notable works include~\cite{graves2021amnesiac,sekhari2021remember,shibata2021learning}. Our method does not put any constraints on the type of optimization to be used while training. We do not train any additional network, in-fact we do not require any prior information related to the training process. In addition, we propose the first Unlearning method that works for both CNN and Vision Transformers. We show the results on different deep learning models, small and large datasets, and demonstrate successful unlearning in face-recognition. 

\subsection{Data Privacy} 
Privacy in machine learning has been extensively studied and various privacy-preserving mechanisms have been presented~\cite{shokri2015privacy,phan2016differential}. The most common assumption in the such privacy protecting frameworks is that the model can freely access the entire training data and algorithms are devised to protect the model from leaking information about the training data. Another privacy setting~\cite{shan2020fawkes,huang2021unlearnable} considers a scenario where the goal is to make the personal data completely unusable for unauthorized deep learning models. The solutions in such a setting are based on the principles of the adversarial attack and defence methods~\cite{moosavi2017universal,shen2019human}. Some privacy settings~\cite{izzo2021approximate,golatkar2021mixed} allow the user to make a request to forget their data from the already trained model. These privacy settings assume having access to all the training data before forgetting. We propose to work in a stricter setting where once the user has made a request for forgetting her data (for example, her face in the face recognition model), the model can not use those samples even for the purpose of network weight manipulation.

\section{Unlearning in Zero-glance Privacy Setting}
\subsection{Zero-glance Privacy Assumptions}
In several use cases, the machine learning model is trained with facial images and personal medical data. Due to the sensitive nature of the data and the time constraints usually set by the data protection regulations (GDPR, CCPA), it may not be possible to use the forget set data even for unlearning purpose. We assume that the user can request for immediate deletion of her data and a time-bound  removal of the information (in network weights) from the already trained model. The immediate removal of requested data leaves us with only the remaining data to perform unlearning. Once the network weights are updated, the model should not have any information corresponding to the forgetting data. Even after being exposed to the forgetting samples, the relearn time should be substantially high to ensure that the model has actually forgotten those samples. 

\subsection{Preliminaries and Objective}
We formulate the unlearning problem in the context of deep networks. Let the complete training dataset consisting of $n$ samples and $K$ total number of classes be $\mathcal{D}_{c} = \{(x_{i}, y_{i})\}_{i=1}^n$ where $x\in \mathcal{X} \subset \mathbb{R}^{d}$ are the inputs and $y\in \mathcal{Y}={1,...,K}$ are the corresponding class labels. If the forget and retain classes are denoted by $\mathcal{Y}_{f}$ and $\mathcal{Y}_{r}$ then $\mathcal{D}_{f} \cup \mathcal{D}_{r} = \mathcal{D}_{c}$, $\mathcal{D}_{f} \cap \mathcal{D}_{r} = \emptyset$. 
Let the deep learning model be represented by the function $f_{\theta}(x): \mathcal{X} \rightarrow \mathcal{Y}$ parameterized by $\theta \in \mathbb{R}^{d}$ used to model the relation $\mathcal{X} \rightarrow \mathcal{Y}$. The weights $\theta$ of the original trained deep network $f_{\theta}$ \footnote{We use the notation $f$ to denote the model in the rest of this paper} are a function of the complete training data $\mathcal{D}_{c}$. Forgetting in zero-glance privacy setting is an algorithm, which gives a new set of weights $\mathcal{\theta}_{D_{r\_sub}}$ by using the trained model $f$ and a subset of retain images $\mathcal{D}_{r\_sub} \subset \mathcal{D}_{r}$ which does not remember the information regarding $D_{f}$ and behaves similarly to a model which has never seen $D_f$ in the parameter and output space.

To achieve unlearning, we first learn a noise matrix $\mathcal{N}$ for each class in $\mathcal{Y}_{f}$ by using the trained model. 
Then we transform the model in such a way that it fails to classify the samples from forget set $\mathcal{D}_{f}$ while maintaining the accuracy for classifying the samples from the retain set $\mathcal{D}_{r}$. This is ensured by using a small subset of samples $\mathcal{D}_{r\_sub}$ drawn from the retain dataset $\mathcal{D}_{r}$.

\section{Error-Maximizing Noise based Unlearning}
Our approach aims to learn a noise matrix for the unlearning class by maximizing the model loss. Such generated noise samples will damage/overwrite the previously learned network weights for the relevant class(es) during the model update and induce unlearning. Error maximizing noise will have high influence to enable parameters updates corresponding to the unlearning class.\par

\subsection{Error-maximizing Noise}
We learn an error-maximizing noise $\mathcal{N}$ of the same size as that of the model input. The goal is to create a correlation between $\mathcal{N}$ and the unlearning class label, $f: \mathcal{N} \rightarrow \mathcal{Y}_{f}, \mathcal{N} \neq \mathcal{X}$. We freeze the weights of the pretrained model during this error maximizing process (see Fig.~\ref{fig:framework}). Given a noise matrix $\mathcal{N}$, initialized randomly with a normal distribution $N(0,1)$, we propose to optimize the error-maximizing noise by solving the following optimization problem:
\begin{equation}
\label{eq1}
\text{arg}\,\min\limits_\mathcal{N} \mathbb{E}_{(\theta)}
 \left[ -\mathcal{L}(f,y) + \lambda\|w_{noise}\|
 \right]
\end{equation}
where, $\mathcal{L}(f,y)$ is the classification loss corresponding to the class to unlearn, $f$ denotes the trained model. The $w_{noise}$ are the parameters of the noise $\mathcal{N}$ (can be interpreted as pixel values in terms of an image) and $\lambda$ is used to manage the trade-off between the two terms. The optimization problem finds the $L_{p}$-norm bounded noise that maximizes the model’s classification loss. In our method, we use a Cross-Entropy loss function $\mathcal{L}$ with $L_{2}$ normalization.

We maximise the error corresponding to the forget class(es) so that this noise is \textit{opposite} to what $D_f$ represents. Using this in the \textit{impair} stage of the UNSIR algorithm erases information related to $D_f$. Overall, it enables efficient unlearning in deep networks. The second term $\lambda\|w_{noise}\|$ in Eq.~\ref{eq1} is proposed to regularize the overall loss by preventing the values in $\mathcal{N}$ from becoming too large. Without this regularization of $\mathcal{N}$, the model will start believing that images with higher values belong to the unlearn class. For multiple classes of data, we learn the noise matrix $\mathcal{N}$ for each class separately. Since the optimization is performed using the model loss w.r.t. the noise matrix, this can be done in an insignificant amount of time. The UNSIR algorithm will be executed only once for both single-class and multi-class unlearning.\par

\subsection{UNSIR: Unlearning with Single Pass Impair and Repair}
We combine the noise matrix along with the samples in $\mathcal{D}_{r\_sub}$ i.e., $\mathcal{D}_{r\_sub} \cup \mathcal{N}$, and train the model for 1 epoch (impair) to induce unlearning. After that we again train (repair) the model for 1 epoch, now on $\mathcal{D}_{r\_sub}$ only. The final model shows excellent performance in unlearning the targeted classes of data and retaining the accuracy on the remaining classes. 

\textbf{Impair.} We train the model on a small subset of data from the original distribution which also contains generated noise. This step is called \textit{impair} as it corrupts those weights in the network which are responsible for recognition of the data in forget class(es). We use a high learning rate and observe that almost always only a single epoch of \textit{impair} is sufficient.

\textbf{Repair.} The \textit{impair} step may sometimes disturb the weights that are responsible for predicting the retain classes. Thus, we \textit{repair} those weights by training the model for a single epoch (on rare occasions, more epochs may be required) on the retain data $\mathcal{D}_{r\_sub}$. The final updated model has high relearn time i.e., it takes substantial number of epochs for the network to relearn the forget samples. This is one of the important criteria for effective unlearning and the proposed method shows good robustness for the same. The overall framework of our unlearning algorithm is shown in Fig.~\ref{fig:framework}.

\begin{table*}[t]
\footnotesize
\centering
\caption{Unlearning on CIFAR-10. \textbf{Original Model:} the model trained on complete dataset $D_c$.~\textbf{Retrain Model:} the model trained on retain set $D_r$.~\textbf{Fine Tune:} the fine tuned model on $D_r$.~\textbf{NegGrad:} the network fine tuned on $D_f$ with negative gradients (Gradient Ascent).~\textbf{Our Method:} the proposed unlearning method.~\textbf{RT:} Relearn Time (RT) is the \textit{number of epochs} taken by model to regain full accuracy on forget set when trained on 500 random samples from $D_c$. A higher value of \textbf{RT} denotes robust erasure of information in the network weights. The accuracy $A_{D_f}$ on the forget set should be close to zero and $A_{D_r}$ should be close to original model's $A_{D_r}$.~\# $\mathcal{Y}_{f}$ denotes the number of unlearning classes.}

\begin{tabular}{c|c|cccccc|cccc}
\hline
\multirow{2}{*}{Model} & \multirow{2}{*}{\# $\mathcal{Y}_{f}$} & \multirow{2}{*}{Metrics} & Original  & Retrain & FineTune & NegGrad & \textbf{Our} & \multicolumn{4}{c}{Relearn Time (RT)}\\
\cline{9-12}
{} & {} & {} & Model & Model & {~\cite{golatkar2020eternal}} & {~\cite{golatkar2020eternal}} &\textbf{Method} 
&Retrain
&FineTune~\cite{golatkar2020eternal}
&NegGrad~\cite{golatkar2020eternal}
&\textbf{Ours}\\
\hline
\multirow{8}{*}{\STAB{\rotatebox[origin=c]{90}{ResNet18}}} & \multirow{2}{*}{1} & $A_{D_r}$ $\uparrow$ & 77.86 & 78.32 & 78.11 & 66.67 & \textbf{71.06 $\pm$1.13} & \multirow{2}{*}{77}  & \multirow{2}{*}{8} & \multirow{2}{*}{54} & \multirow{2}{*}{90}\\

{} & {} & $A_{D_f}$ $\downarrow$ & 81.01 & 0 & 24.55 & 7.44 & \textbf{0 $\pm$0} & {} & {} & {} & {} \\
\cline{2-12}
 & \multirow{2}{*}{2} & $A_{D_r}$ $\uparrow$ & 78.00 & 79.15 & 79.53 & 72.12 & \textbf{73.61 $\pm$0.51} & \multirow{2}{*}{$>100$} & \multirow{2}{*}{10} & \multirow{2}{*}{7}  & \multirow{2}{*}{\textbf{$>100$}}\\
{} & {} & $A_{D_f}$ $\downarrow$ & 78.65 & 0 & 31.59 & 0.05 & \textbf{0 $\pm$0} & {} & {} & {} & {} \\
\cline{2-12}
 & \multirow{2}{*}{4} & $A_{D_r}$ $\uparrow$ & 81.42 & 85.88 & 85.49 & 54.84 & \textbf{76.63 $\pm$0.89} & \multirow{2}{*}{$>100$} & \multirow{2}{*}{13} & \multirow{2}{*}{18} & \multirow{2}{*}{\textbf{$>100$}}\\
{} & {} & $A_{D_f}$ $\downarrow$ & 73.45 & 0 & 41.45 & 0.02 & \textbf{0 $\pm$0} & {} & {} & {} & {}\\
\cline{2-12}
 & \multirow{2}{*}{7} & $A_{D_r}$ $\uparrow$ & 79.36 & 91.39 & 79.27 & 31.87 & \textbf{82.86 $\pm$1.42} &  \multirow{2}{*}{$>100$} & \multirow{2}{*}{0} & \multirow{2}{*}{$>100$} & \multirow{2}{*}{\textbf{$>100$}}\\
{} & {} & $A_{D_f}$ $\downarrow$ & 77.58 & 0 & 77.71 & 0.19 & \textbf{0 $\pm$0} & {} & {} & {} & {}\\
\hline
\multirow{8}{*}{\STAB{\rotatebox[origin=c]{90}{AllCNN}}} & \multirow{2}{*}{1} & $A_{D_r}$ $\uparrow$ & 82.64 & 85.90 & 85.01 & 42.39 & \textbf{73.90 $\pm$1.09} & \multirow{2}{*}{$>100$} & \multirow{2}{*}{12} & \multirow{2}{*}{18} & \multirow{2}{*}{\textbf{$>100$}} \\
{} & {} & $A_{D_f}$ $\downarrow$ & 91.02 & 0 & 31.07 & 13.47 & \textbf{0 $\pm$0} & {} & {} & {} & {} \\
\cline{2-12}
 & \multirow{2}{*}{2} & $A_{D_r}$ $\uparrow$ & 84.27 & 85.21 & 86.45 & 44.25 & \textbf{80.76 $\pm$1.33} &  \multirow{2}{*}{$>100$}  & \multirow{2}{*}{6} & \multirow{2}{*}{7} & \multirow{2}{*}{74}\\
{} & {} & $A_{D_f}$ $\downarrow$ & 79.74 & 0 & 35.45 & 2.20 & \textbf{0 $\pm$0} & {} & {} & {} & {}\\
\cline{2-12}
 & \multirow{2}{*}{4} & $A_{D_r}$ $\uparrow$ & 87.06 & 91.86 & 91.75 & 22.00 & \textbf{80.21 $\pm$0.73} &  \multirow{2}{*}{$>100$} & \multirow{2}{*}{4} & \multirow{2}{*}{24} & \multirow{2}{*}{\textbf{$>100$}}\\
{} & {} & $A_{D_f}$ $\downarrow$ & 78.00 & 0 & 53.66 & 1.44 & \textbf{0 $\pm$0} & {} & {} & {} & {}  \\
\cline{2-12}
 & \multirow{2}{*}{7} & $A_{D_r}$ $\uparrow$ & 83.30 & 94.47 & 83.29 & 22.96 & \textbf{85.21 $\pm$1.44} & \multirow{2}{*}{$>100$} & \multirow{2}{*}{0} & \multirow{2}{*}{$>100$} & \multirow{2}{*}{\textbf{$>100$}}\\
{} & {} & $A_{D_f}$ $\downarrow$ & 83.31 & 0 & 83.28 & 0 & \textbf{0 $\pm$0} & {} & {} & {} & {}\\
\hline
\end{tabular}
\label{CIFAR-10}
\end{table*}
 

\section{Experiments and Results}
\label{experiments}
We show the performance of our proposed method for unlearning single and multiples classes of data across a variety of settings. We use different types of deep networks ResNet18~\cite{he2016deep}, AllCNN~\cite{springenberg2014striving}, MobileNetv2~\cite{sandler2018mobilenetv2} and Vision Transformers~\cite{dosovitskiy2020image} for evaluation and empirically demonstrate the applicability of our method across these different networks. The experiments are conducted for network trained from scratch as well as pretrained models fine-tuned on specific datasets.
The unlearning method is analyzed over CIFAR-10~\cite{krizhevsky2009learning}, CIFAR-100~\cite{krizhevsky2009learning} and VGGFace-100 (100 face IDs collected from the VGGFaces2~\cite{cao2018vggface2}). Results on these variety of models and datasets demonstrate the wide applicability of our method.\par

The experimental results are reported with a single step (1 epoch) of \textit{impair} and a single step of \textit{repair}. Additional fine-tuning could be done, however, we focus on such a setting (single-shot) to demonstrate the efficacy of our method under a uniform setup. All the models learned from scratch have been trained for 40 epochs, and the pretrained models have been fine-tuned for 5 epochs. We observe that $\lambda = 0.1$ in Eq.~\ref{eq1} works quiet well across various tasks, and thus keep it fixed at 0.1 for all the experiments.\par


\begin{table*}[]
\footnotesize
\centering
\caption{Comparison of our method with a single class Fisher Forgetting~\cite{golatkar2020eternal} method on CIFAR-10. Fisher achieves forgetting but fails to maintain the accuracy on the retained dataset.}
\begin{tabular}{*7c}
\hline
\multirow{2}{*}{} &  \multicolumn{2}{c}{Initial Accuracy} & \multicolumn{2}{c}{Fisher Forgetting~\cite{golatkar2020eternal}} & \multicolumn{2}{c}{\textbf{Our Method}}\\
\hline
{Model} & $A_{D_f}$ $\downarrow$ & $A_{D_r}$ $\uparrow$ & $A_{D_f}$ $\downarrow$ & $A_{D_r}$ $\uparrow$ & \textbf{$A_{D_f}$} $\downarrow$& \textbf{$A_{D_r}$} $\uparrow$\\
\hline
ResNet18 &  81.01 & 77.86 & 0 & 10.85 & \textbf{0} & \textbf{71.06}\\
AllCNN & 91.02 & 82.64 & 0 & 7.61 & \textbf{0} & \textbf{73.90}\\
\hline
\end{tabular}
\label{fisher}
\end{table*}

\begin{table}[]
\centering
\label{CIFAR-100}
\caption{Unlearning on CIFAR-100. The models are pretrained on ImageNet and fine tuned for CIFAR-100} 
\centering
\begin{tabular}{c|c|cccc}
\hline
\multirow{2}{*}{Model} & \multirow{2}{*}{\# $\mathcal{Y}_{f}$} & \multirow{2}{*}{Metrics} & Original  & Retrain & \textbf{Our} \\
{} & {} & {} & Model & Model &\textbf{Method}\\
\hline
\multirow{8}{*}{\STAB{\rotatebox[origin=c]{90}{ResNet18}}} & \multirow{2}{*}{1} & $A_{D_r}$ $\uparrow$ & 78.68 & 78.37 & \textbf{75.36}\\
 & {} & $A_{D_f}$ $\downarrow$ & 83.00 & 0 & \textbf{0} \\
\cline{2-6}

 & \multirow{2}{*}{20} & $A_{D_r}$ $\uparrow$ & 77.88 & 79.73 & \textbf{75.38}\\
 & {} & $A_{D_f}$ $\downarrow$ & 82.84 & 0 & \textbf{0} \\
\cline{2-6}
 & \multirow{2}{*}{40} & $A_{D_r}$ $\uparrow$ & 78.31 & 82.65 & \textbf{78.85}\\
 & {} & $A_{D_f}$ $\downarrow$ & 79.78 & 0 & \textbf{0}\\
\cline{2-6}
 & \multirow{2}{*}{60} & $A_{D_r}$ $\uparrow$ & 76.96 & 83.62 & \textbf{75.51}\\
 & {} & $A_{D_f}$ $\downarrow$ & 80.31 & 0 & \textbf{0.47}\\
\hline
\multirow{8}{*}{\STAB{\rotatebox[origin=c]{90}{MobileNetv2}}} & \multirow{2}{*}{1} & $A_{D_r}$ $\uparrow$ & 77.43 & 78 & \textbf{75.76}\\
 & {} & $A_{D_f}$ $\downarrow$ & 90 & 0 & \textbf{0}\\
\cline{2-6}
 & \multirow{2}{*}{20} & $A_{D_r}$ $\uparrow$ & 76.47 & 77 & \textbf{76.27}\\
 & {} & $A_{D_f}$ $\downarrow$ & 81.70 & 0 & \textbf{0}\\
\cline{2-6}
 & \multirow{2}{*}{40} & $A_{D_r}$ $\uparrow$ & 76.93 & 80.24 & \textbf{77.66}\\
 & {} & $A_{D_f}$ $\downarrow$ & 78.56 & 0 & \textbf{0.02}\\
\cline{2-6}
 & \multirow{2}{*}{60} & $A_{D_r}$ $\uparrow$ & 76.17 & 79.37 & \textbf{68.57}\\
 & {} & $A_{D_f}$ $\downarrow$ & 78.56 & 0 & \textbf{1.22}\\
\hline
\end{tabular}
\label{CIFAR-100}
\end{table}

\begin{table}[]
\centering
\label{VGGFace-100}
\caption{Unlearning on VGGFace-100. The models are pretrained on ImageNet and fine tuned for VGGFace-100}
\begin{tabular}{c|c|cccc}
\hline
\multirow{2}{*}{Model} & \multirow{2}{*}{\# $\mathcal{Y}_{f}$} & \multirow{2}{*}{Metrics} & Original  & Retrain & \textbf{Our} \\
{} & {} & {} & Model & Model &\textbf{Method}\\
\hline
\multirow{8}{*}{\STAB{\rotatebox[origin=c]{90}{ResNet18}}} & \multirow{2}{*}{1} & $A_{D_r}$ $\uparrow$ & 80.63 & 80.42 & \textbf{72.79}\\
{} & {} & $A_{D_f}$ $\downarrow$ & 94.00 & 0 & \textbf{3.00} \\
\cline{2-6}
 & \multirow{2}{*}{20} & $A_{D_r}$ $\uparrow$ & 81.15 & 69.96 & \textbf{73.26}\\
{} & {} & $A_{D_f}$ $\downarrow$ & 78.45 & 0 & \textbf{0.15}\\
\cline{2-6}
 & \multirow{2}{*}{40} & $A_{D_r}$ $\uparrow$ & 81.31 & 82.74 & \textbf{78.66}\\
{} & {} & $A_{D_f}$ $\downarrow$ & 79.18 & 0 & \textbf{6.71}\\
\cline{2-6}
 & \multirow{2}{*}{60} & $A_{D_r}$ $\uparrow$ & 81.30 & 82.66 & \textbf{79.16}\\
{} & {} & $A_{D_f}$ $\downarrow$ & 80.03 & 0 & \textbf{8.62}\\
\hline
\multirow{8}{*}{\STAB{\rotatebox[origin=c]{90}{ViT}}}& \multirow{2}{*}{1} & $A_{D_r}$ $\uparrow$ & 91.53 & 92.45 & \textbf{82.90}\\
& {} & $A_{D_f}$ $\downarrow$ & 74.22 & 0 & \textbf{4.81}\\
\cline{2-6}
 & \multirow{2}{*}{20} & $A_{D_r}$ $\uparrow$ & 91.52 & 93.70 & \textbf{85.21}\\
 & {} & $A_{D_f}$ $\downarrow$ & 91.30 & 0 & \textbf{26.00}\\
\cline{2-6}
 & \multirow{2}{*}{40} & $A_{D_r}$ $\uparrow$ & 92.10 & 94.13 & \textbf{85.33}\\
 & {} & $A_{D_f}$ $\downarrow$ & 90.55 & 0 & \textbf{25.10}\\
\cline{2-6}
 & \multirow{2}{*}{60} & $A_{D_r}$ $\uparrow$ & 90.97 & 93.35 & \textbf{87.82}\\
 & {} & $A_{D_f}$ $\downarrow$ & 91.82 & 0 & \textbf{8.48}\\
\hline
\end{tabular}
\label{VGGFace-100}
\end{table}

\subsection{Evaluation Metrics}
In the literature~\cite{golatkar2020eternal,golatkar2020forgetting,golatkar2021mixed,graves2021amnesiac} several metrics have been defined to measure the overall performance of an unlearning method. These metrics attempt to determine the amount of information remaining in the network about the unlearn/forget data. In our analysis, we use the following metrics: 

\textbf{Accuracy on forget set ($A_{D_{f}}$):} Should be close to zero.\par
\textbf{Accuracy on retain set ($A_{D_{r}}$):} Should be close to the performance of original model.\par

\textbf{Relearn time ($RT$):} Relearn time is a good proxy to measure the amount of information remaining in the model about the unlearning data. If a model regains the performance on the unlearn data very quickly with only few steps of retraining, it is highly likely that some information regarding the unlearn data is still present in the model. We measure the relearn time as the number of epochs it takes for the unlearned model to reach the source model's accuracy, with the model being trained on 500 random samples from the training set in each epoch.\par

\textbf{Weight distance:} The distance between individual layers of the original model and the unlearned model gives additional insights about the amount of information remaining in the network about the forget data. A comparative analysis with the retrained model would validate the robustness of the unlearning method.\par
\textbf{Prediction distribution on forget class:} We analyze the distribution of the predictions for different samples in the forget class(es) of data in the unlearned model. Presence of any specific observable patterns like repeatedly predicting a single retain class may indicate risk of information exposure.\par

Additionally, a high similarity with the prediction distribution of the retrain model would indicate robustness in the unlearning method to information exposure of the forget class. A recent work~\cite{rezaei2021difficulty} has reported the shortcomings of membership inference attacks on deep networks. Thus, we avoid using them to keep the analysis more consistent and reliable. It is to be noted that a comprehensive method of evaluating the exposure/leakage of private data in a deep model is a difficult task~\cite{graves2021amnesiac}, and we are not aware of any method claiming to do so.

\subsection{Models}
In CIFAR-10, we trained~\textit{ResNet18} and \textit{AllCNN} from scratch and used the proposed method to unlearn a single class and multiple classes (2 classes, 4 classes, and 7 classes) from the model. Without loss of generality, we use class 0 for single class unlearning, and a random manual selection of class subsets for multi-class unlearning. For example, in 2-class unlearning we unlearn class 1-2, in 4-class unlearning we unlearn classes 3-6, in 7-class unlearning we unlearn classes 3-9. In CIFAR-100, we use pretrained \textit{ResNet18} and \textit{MobileNetv2}. The unlearning is performed for 1 class (class 0), and 20 and 40 randomly sampled classes. In the later part, we also demonstrate unlearning on VGGFace-100 using pretrained \textit{ResNet18} and \textit{Vision Transformer}. The unlearning is performed for 1-faceID, 20-faceID, 40-faceID, and 60-faceID, respectively.

\subsection{Baseline Unlearning Methods}
We primarily use the following baseline methods: (i) fine-tuning on the retain set i.e., catastrophic forgetting (\textit{FineTune}), (ii) gradient ascent on the forget class (\textit{NegGrad}). The comparative results are shown in Table~\ref{CIFAR-10}. We also run \textit{Fisher Forgetting}~\cite{golatkar2020eternal} and show the results in Table~\ref{fisher}. We present the results in two models for 1-class forgetting as the Fisher method is computationally very expensive. We did not use methods like removing the corresponding class from the final output as it does not remove any information from the model itself. Simply removing the final layer class might also lead to Streisand effect i.e., the information we are trying to hide may become even more prominent.

\subsection{Experimental Settings} 
The experiments are conducted on a NVIDIA Tesla-V100 (32GB) GPU. The settings for individual dataset are given below:\par

\textbf{CIFAR-10:} The error-maximizing noise is learned for a single batch size and 20 copies of this noise are used for the noise data set. A batch size of 256 is used for all the datasets. The retain set ($D_{r}$) is created by collecting 1000 samples of each retain class. The learning rate of 0.02 is used for impair step, where 1 epoch (one shot of damage) is trained done using the mix of retain sub-samples and noise. The learning rate in repair step is 0.01, where 1 epoch (1 shot of healing) is trained on the retain sub-samples.\par

\textbf{CIFAR-100:} 
Same as in CIFAR-10, the error-maximizing noise is learned for a single batch size and 20 copies of this noise are used for the noise data set. The retain set consists of 50 samples collected from each retain class. For pre-trained ResNet18, the learning rate in the impair step is set to 0.01 for the last layer and 0.0001 for the remaining layers. Likewise, in the repair step, the learning rate is set to 0.005 for the last layer and 0.0001 for rest of the layers. In the AllCNN model, the learning rate for impair and repair steps are 0.02 and 0.01, respectively.\par

\textbf{VGGFace-100:} A batch of the noise matrix is learned and copied 15 times to create the noise data set. The retain set consists of 100 samples of each retain class. For ResNet18, the learning rate in impair and repair steps are 0.01 and 0.001, respectively. For Vision Transformer model, the learning rate for impair and repair steps are 0.0001 and 0.00002, respectively. We also run a Fisher Forgetting model as presented in~\cite{golatkar2020eternal} which is similar to a targeted noise addition based approach.


\begin{figure*}[]
\centering
\begin{minipage}{\linewidth}
  \centering
  \begin{tabular}{cccc}
  \includegraphics[width=.23\linewidth]{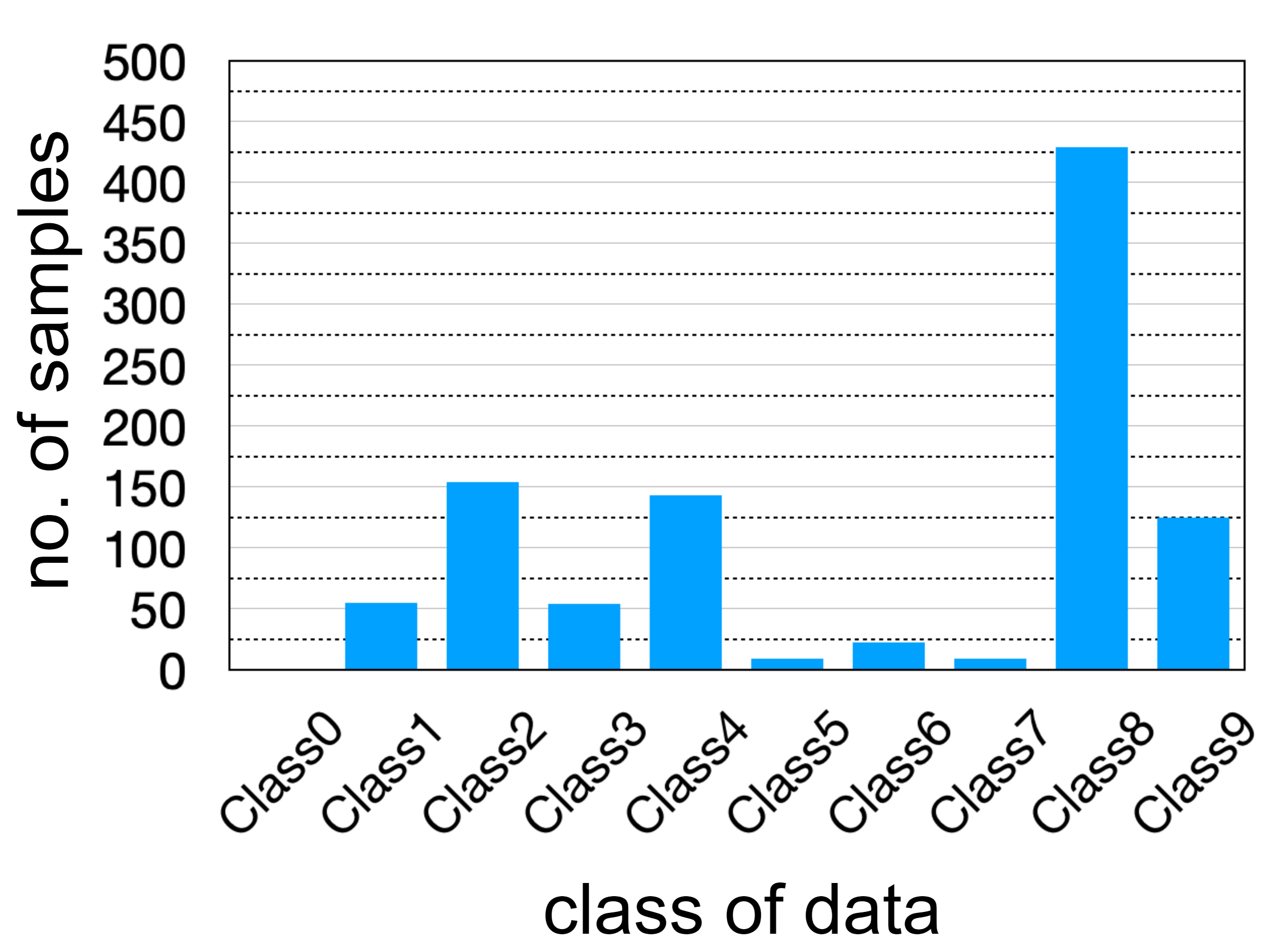} & 
\includegraphics[width=.23\linewidth]{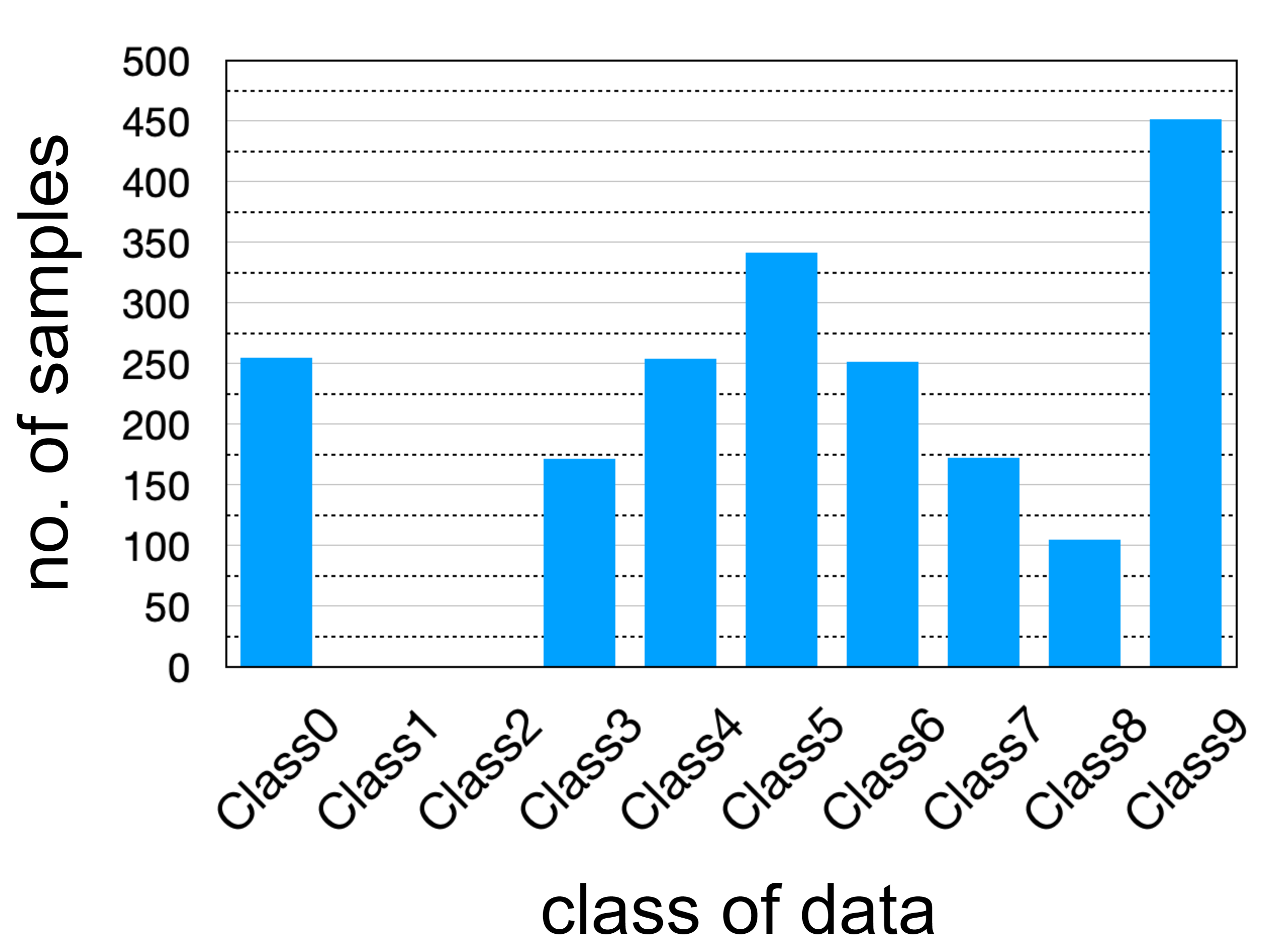} &
  \includegraphics[width=.23\linewidth]{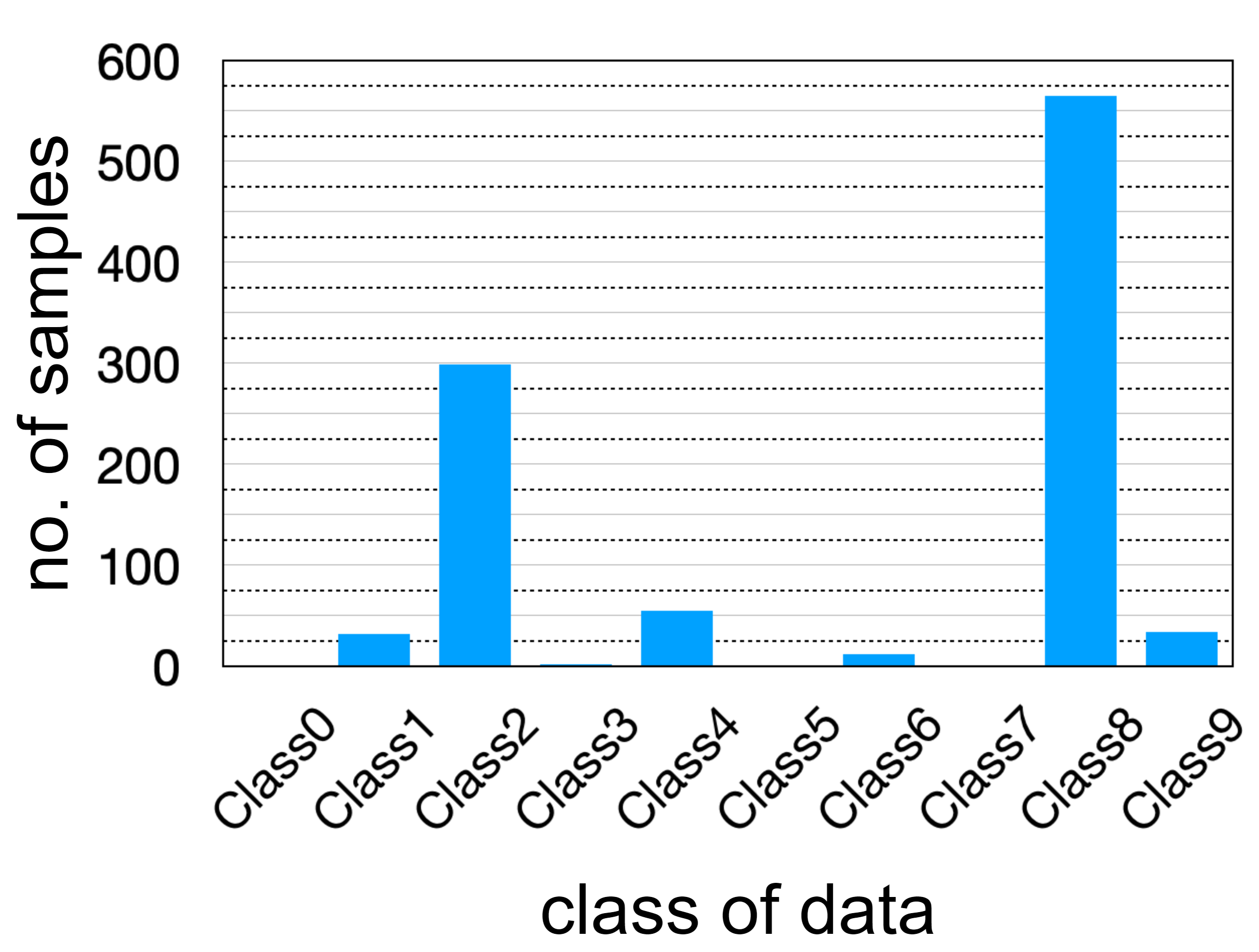} &
  \includegraphics[width=.23\linewidth]{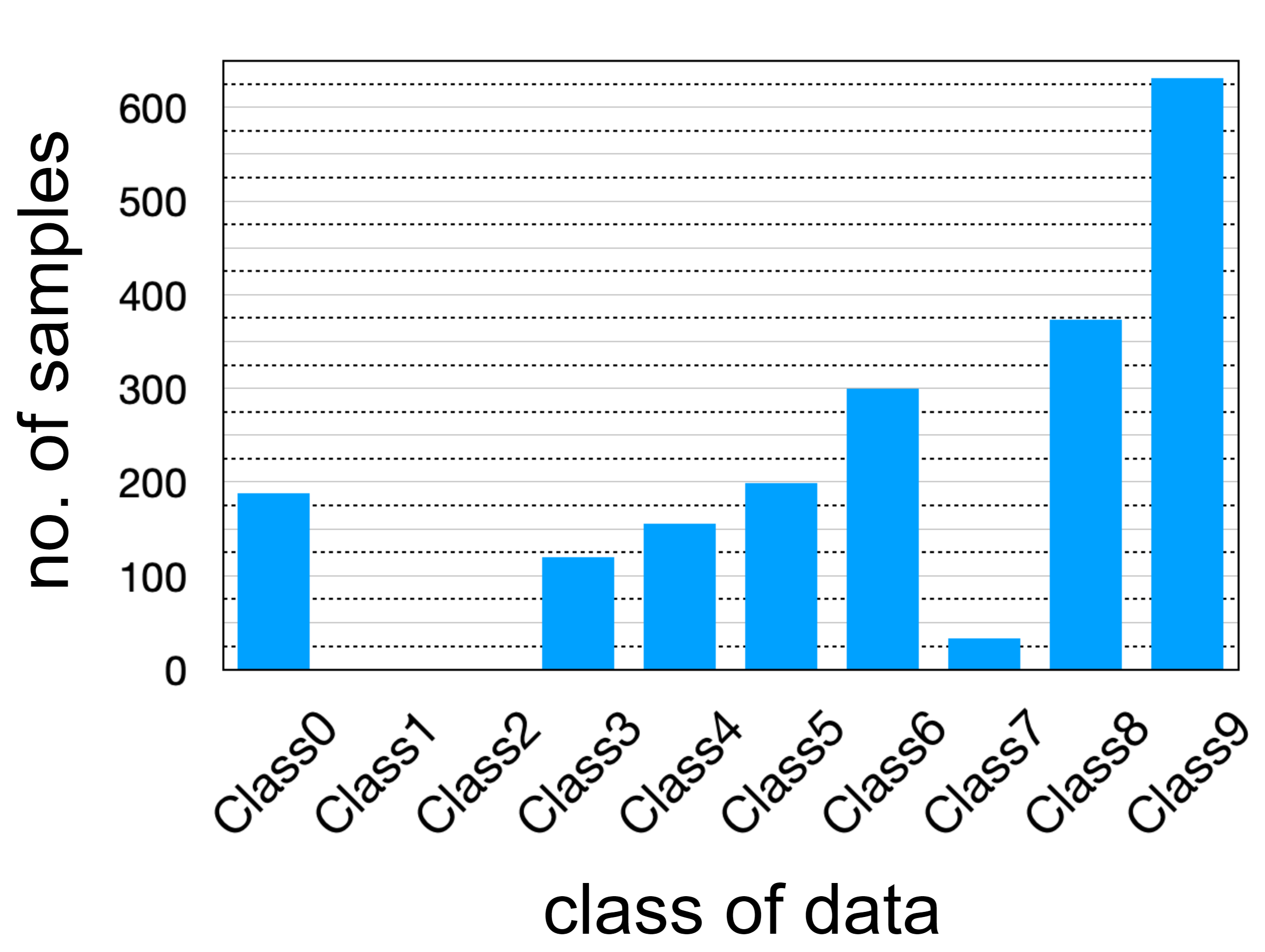}\\
  \scriptsize{a. 1-C unlearning (AllCNN)} &
  \scriptsize{b. 2-C unlearning (AllCNN)}&
  \scriptsize{c. 1-C unlearning (ResNet18)}&
  \scriptsize{d. 2-C unlearning (ResNet18)}\\
  \end{tabular}
\end{minipage}
\caption{Prediction distribution of the unlearned model on forget class of data. Our method gives randomized response to the input query of the forget class of data.}
\label{fig:output_dist}
\end{figure*}

\begin{figure}[]
\centering
\includegraphics[width=0.7\linewidth]{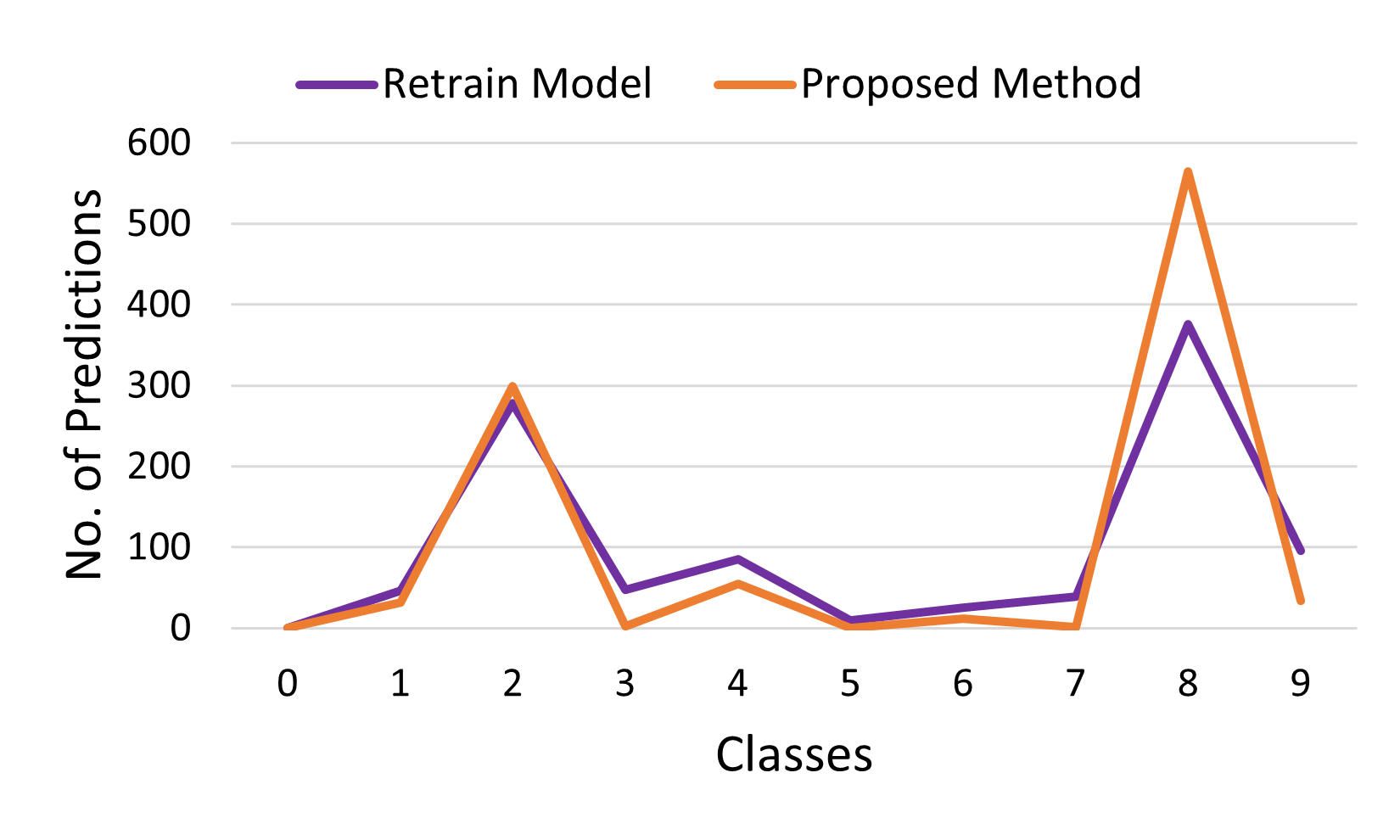}
 \caption{Prediction distribution (AllCNN) of the retrained model (left) and proposed method (right) for the forget class. We can see the distributions are similar.}
 \label{fig:prediction_dist_compare}
\end{figure}

\begin{figure*}[]
\centering
\begin{minipage}{\linewidth}
  \centering
  \begin{tabular}{cccc}
  \includegraphics[width=.23\linewidth]{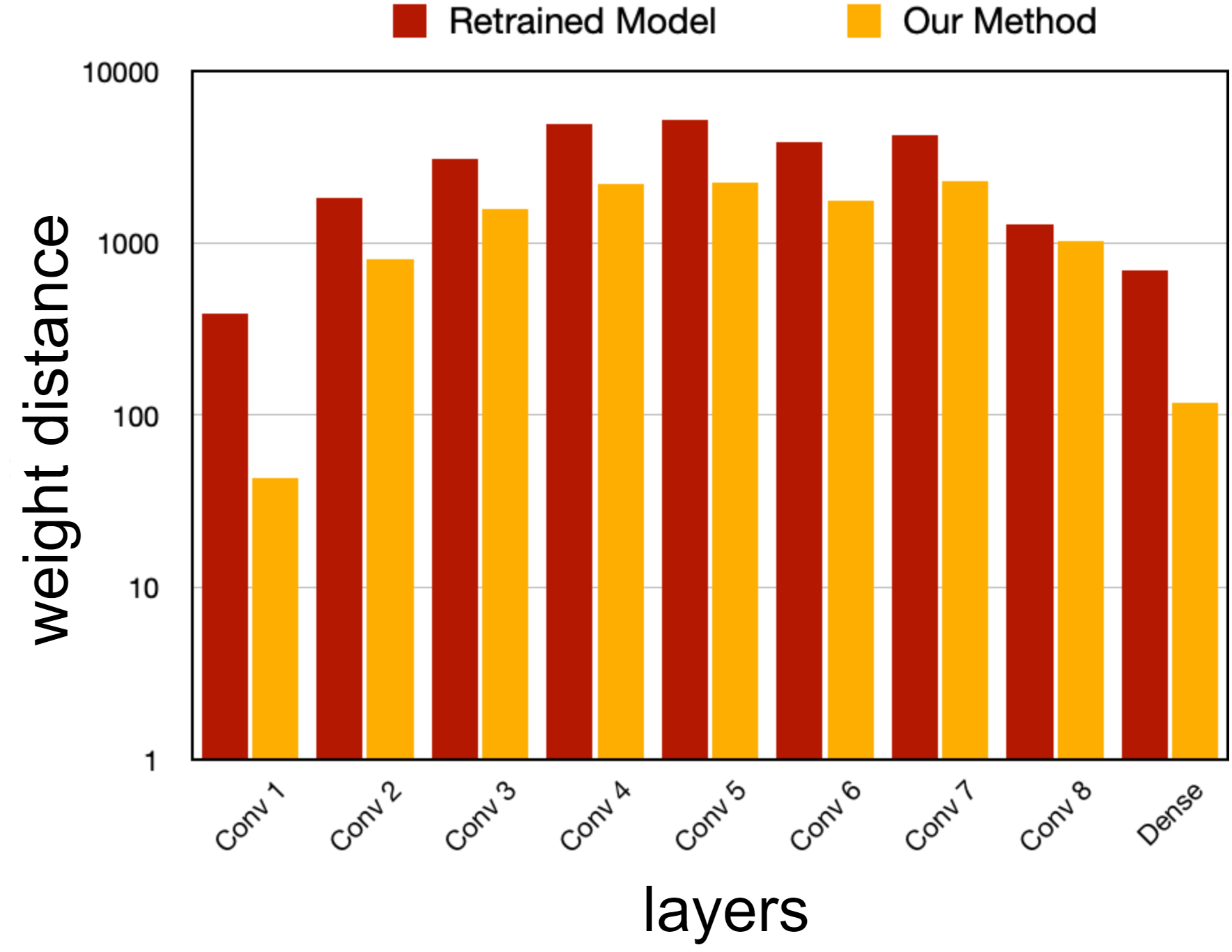}&
  \includegraphics[width=.23\linewidth]{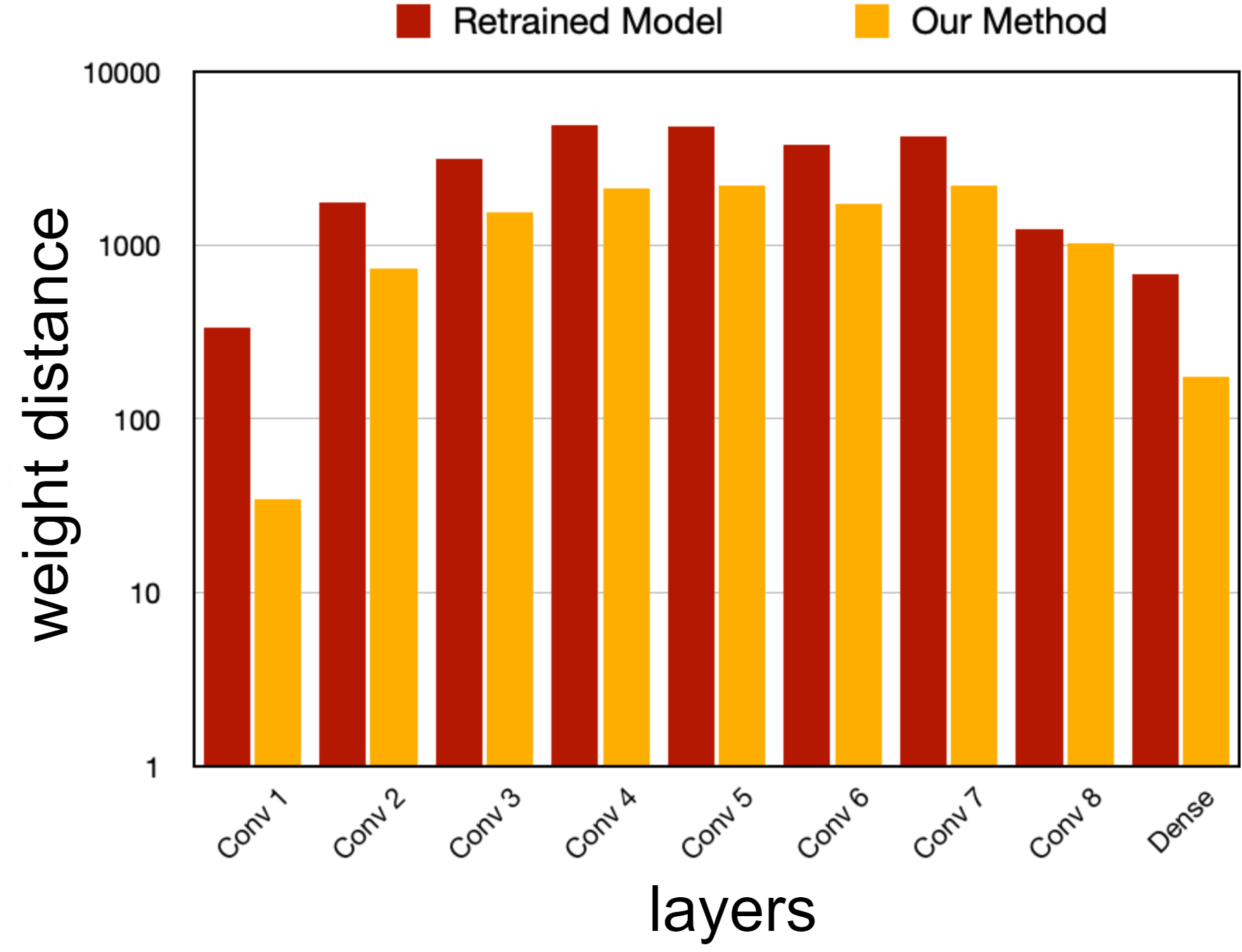}&
  \includegraphics[width=.23\linewidth]{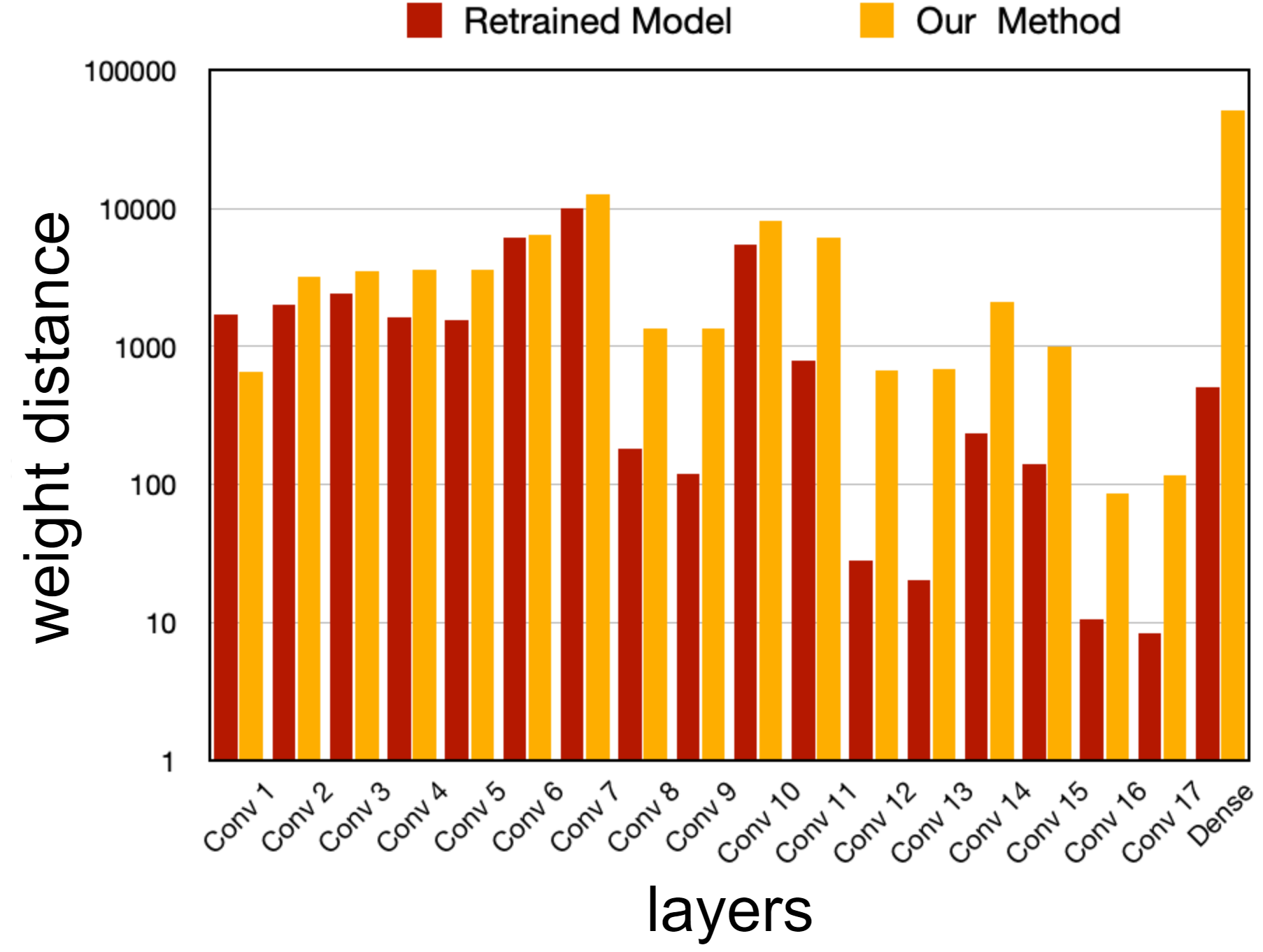} &
  \includegraphics[width=.23\linewidth]{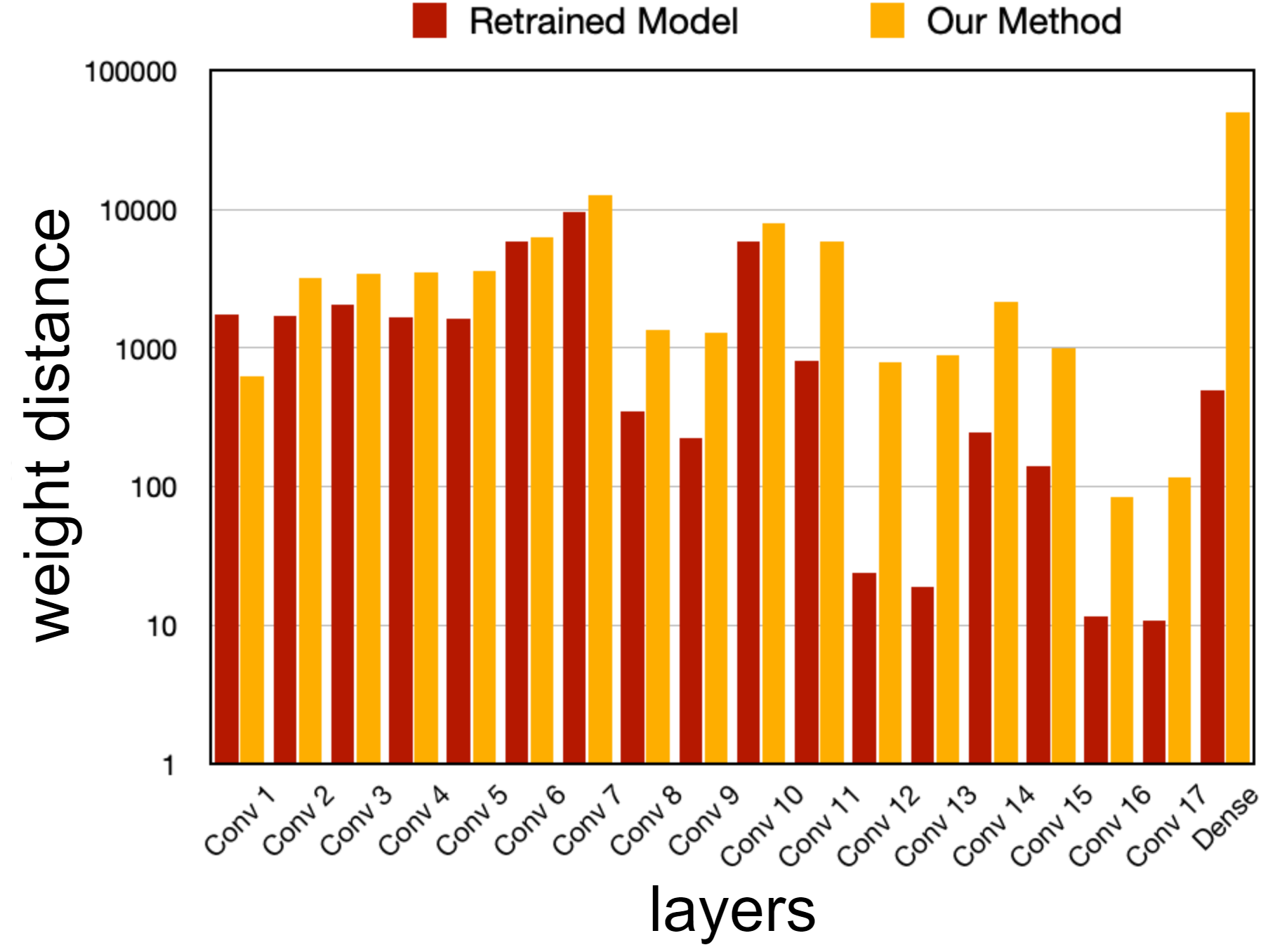}\\
  \scriptsize{a. 1-C unlearning (AllCNN)} &
  \scriptsize{b. 2-C unlearning (AllCNN)}&
  \scriptsize{c. 1-C unlearning (ResNet18)}&
  \scriptsize{d. 2-C unlearning (ResNet18)}\\
  \end{tabular}
\end{minipage}
\caption{Layer-wise weight distance between the unlearned models (retrain model, our model) and the original model. The values are presented on a log scale. Our method obtains comparable or higher weight distances in comparison to the retrain model.}
\label{fig:layer_diff}
\end{figure*}

\subsection{Results}
Our results are compared with three baseline unlearning methods: Retrain Model, FineTune~\cite{golatkar2020eternal}, and NegGrad~\cite{golatkar2020eternal}. We compare the single-class unlearning results with an existing Fisher forgetting method in Table~\ref{fisher}. Due to poor results of FineTune, NegGrad, and Fisher forgetting~\cite{golatkar2020eternal} in CIFAR-10, we compare our results only with the Retrain Model in the subsequent experiments. 

\subsubsection{Single Class Unlearning} Table~\ref{CIFAR-10} and Table~\ref{CIFAR-100} show that our model is able to erase the information with respect to a particular class and unlearn in a single shot of impair and repair. Table~\ref{CIFAR-10} shows the retain and forget set accuracy after unlearning along with average standard deviation after 3 runs. We obtain superior accuracy in retain set ($D_{r}$) and forget set ($D_{f}$) over the existing methods; like in case of ResNet18 and CIFAR-10, we preserve 71.06\% of accuracy on $D_{r}$ from an initial 77.86\% while degrading the performance on $D_{f}$ significantly (0\% from an initial 81.01\%). The relearn time (RT) is much higher for our method in comparison to the baseline methods (for example, $>$100 vs 12, 18 in case of AllCNN, CIFAR-10). This shows the capability of our method to enforce robust unlearning. From Table~\ref{fisher}, we observe that our method is far superior to Fisher forgetting as well. Fisher Forgetting is able to preserve only 10.85\% accuracy in ResNet on $D_{r}$ on CIFAR-10.

\subsubsection{Multiple Class Unlearning}
Our method shows excellent unlearning result for multi-class unlearning. We observe that as the number of class to unlearn increases, the repair step becomes more effective and leads to performance closer to the original model on $D_{r}$. The experiments are done with pretrained ResNet18 on CIFAR-100. After unlearning 20 classes we retain 75.38\% accuracy compared to an initial 77.88\% on retain set. The FineTune and gradient ascent (NegGrad) methods either lose performance on $D_{r}$ or their performance on $D_{f}$ is much higher than expected. For example, in case of 4-class unlearning on CIFAR-10, FineTune retains decent accuracy on $D_{r}$ but it fails to unlearn $D_{f}$ properly. It preserves 53.66\% accuracy on the forget classes vs 0\% preserved by our method. The NegGrad appears to unlearn the forget classes properly but its performance on $D_r$ takes a hit. It obtains 22\% retain set accuracy vs 80.21\% accuracy obtained by our method. In addition, our method significantly outperforms both FineTune and NegGrad in relearn time (RT). This suggests that much of the information about $D_f$ is still present in the unlearned model which is not desirable. For example, in case of 2-class unlearning on CIFAR-10+ResNet-18, NegGrad achieves a decent 72.12\% on $D_r$ and 0.05\% on $D_f$. But the model relearns in 7 epochs compared to our method's RT of 100 epochs. Thus, our method shows excellent overall unlearning results as reported in Table~\ref{CIFAR-10}, Table~\ref{CIFAR-100}, Table~\ref{VGGFace-100} for multi-class unlearning.

\begin{figure}[]
\centering
\begin{minipage}{\linewidth}
  \centering
  \begin{tabular}{ccccc}
    \includegraphics[width=0.15\linewidth]{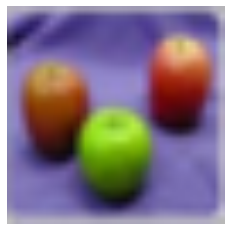} &
    \includegraphics[width=0.15\linewidth]{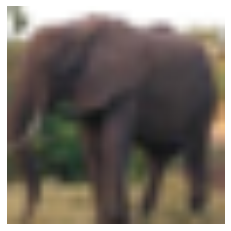} &
    \includegraphics[width=0.15\linewidth]{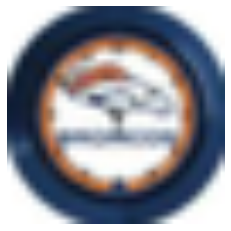} &
    \includegraphics[width=0.15\linewidth]{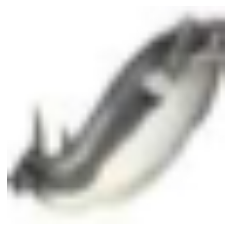} &
    \includegraphics[width=0.15\linewidth]{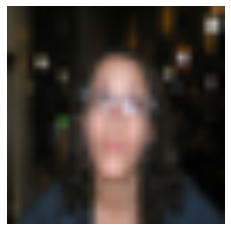} \\
 \multicolumn{5}{c}{\scriptsize{a. Input}}
  \\
      \includegraphics[width=0.15\textwidth]{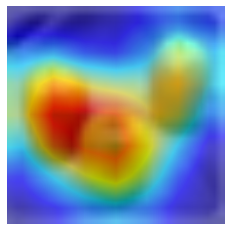}&
    \includegraphics[width=0.15\textwidth]{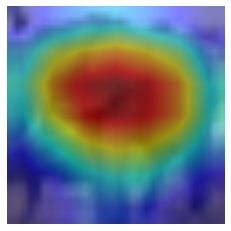}&
    \includegraphics[width=0.15\textwidth]{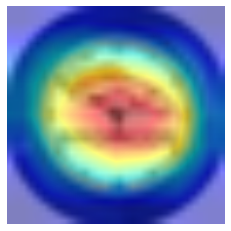}&
    \includegraphics[width=0.15\textwidth]{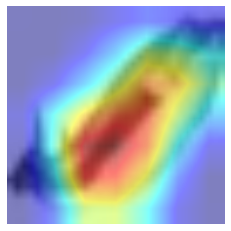}&
    \includegraphics[width=0.15\textwidth]{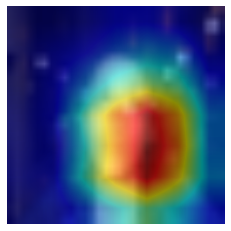}\\
 \multicolumn{5}{c}{\scriptsize{b. Original Model}}
  \\
    \includegraphics[width=0.15\textwidth]{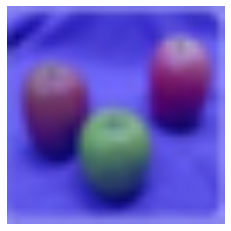}&
    \includegraphics[width=0.15\textwidth]{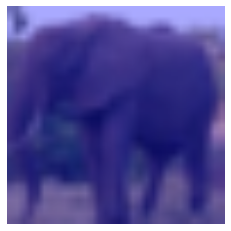}&
    \includegraphics[width=0.15\textwidth]{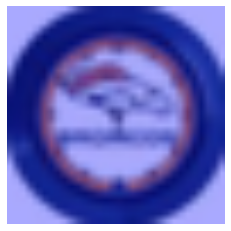}&
    \includegraphics[width=0.15\textwidth]{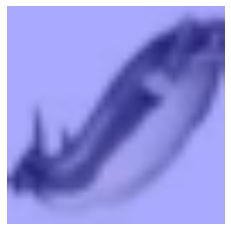}&
    \includegraphics[width=0.15\textwidth]{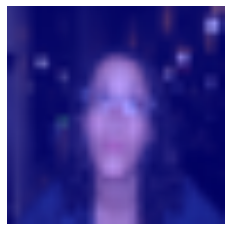}\\
 \multicolumn{5}{c}{\scriptsize{c. Unlearned Model}}
  \\
  \end{tabular}
\end{minipage}
\caption{GradCAM visualization of ResNet18 on CIFAR-100. The first column depicts visualization in 1-class unlearning and remaining columns depict the visualization in 4-classes unlearning.}
\label{fig:gradcam}
\end{figure}

\subsubsection{Unlearning in Face Recognition}
Facial images are difficult to differentiate from each other for a model and is one of the most challenging unlearning task. The results on VGGFace-100 is obtained using ResNet and pretrained Vision Transformer (ViT) and reported in Table~\ref{VGGFace-100}. We report the unlearning performance on $D_r$ and $D_f$ after forgetting 1 class, 20 classes, 40 classes and 60 classes. As the Vision Transformer model is obtained with few epochs (5 epochs) of fine-tuning, the relearn time is expected to be low as well. Therefore, we do not present the analysis corresponding to the relearn time. Our method achieves good retain as well as forgetting accuracy. Like for 1 class forgetting on ResNet18, our method preserves 72.29\% accuracy on $D_r$ compared to an initial 80.63\% and degrades the performance on $D_f$ to 3\%. In case of 60 classes forgetting on pretrained ViT, our method preserves 87.82\% accuracy (initial accuracy: 90.97\%) on $D_r$ and 8.48\% (initial accuracy: 91.82\%) on the $D_f$. This showcases the wide applicability of our method.

\subsection{Prediction Distribution for the Forget Class of Data}
We plot the graph of the prediction class outcomes of the unlearned model for the forget class of data. For example, Fig.~\ref{fig:output_dist} (a) depicts the prediction outcomes of an unlearned ResNet18 model (\textit{forget class = class0}) for the samples from \textit{class0}. The prediction outcomes for 2-classes unlearned model (\textit{forget class = class1, class2}) is also shown in~Fig.~\ref{fig:output_dist} (b). Here we can check whether our unlearned model predicts a specific class(es) for all the forget set of data (Streisand effect~\cite{golatkar2020eternal}), because this could lead to a potential vulnerability to adversarial attacks. We observe in Fig.~\ref{fig:output_dist} that all the predictions for the forget class of data are randomly distributed across the remaining retain classes. Our unlearned model is unable to confidently correlate the forget data with any specific retain class. This shows that our method has actually erased the information related to the unlearn class of data.\par
We also compare the predictions of the retrained model (gold model) and the proposed method in Fig.~\ref{fig:prediction_dist_compare}. It can be observed that the output distribution in both models are very similar. This further shows the robustness of our method.

\section{Analysis}
\subsection{Layer-wise Distance between the Network Weights} 
The layer-wise distance between the original and unlearned models help in understanding the effect of unlearning at each layer. The weight difference should be comparable to the retrain model as a lower distance indicates ineffective unlearning and a much higher distance may point to Streisand effect and possible information leaks.
We compare the weight distance in the (i) retrained model, and (ii) proposed method for AllCNN and ResNet18 in~Fig.~\ref{fig:layer_diff}. We notice that the weight differences of the proposed method with respect to the original model show a similar trend to that of the retrain model.

\subsection{Visualizing the Unlearning in Models}
We use GradCAM~\cite{selvaraju2017grad} to visualize the area of focus in the model (ResNet18) for images in the unlearn class. Fig.~\ref{fig:gradcam} depicts where the model focuses before and after applying our method for unlearning 1-class and 20-classes, respectively. As expected, after applying our method, the model is unable to focus on the relevant areas, indicating that the network weights no longer contain information related to those unlearn classes.

\subsection{Efficiency}
\begin{figure}[]
    \centering
    \includegraphics[width=.5\linewidth]{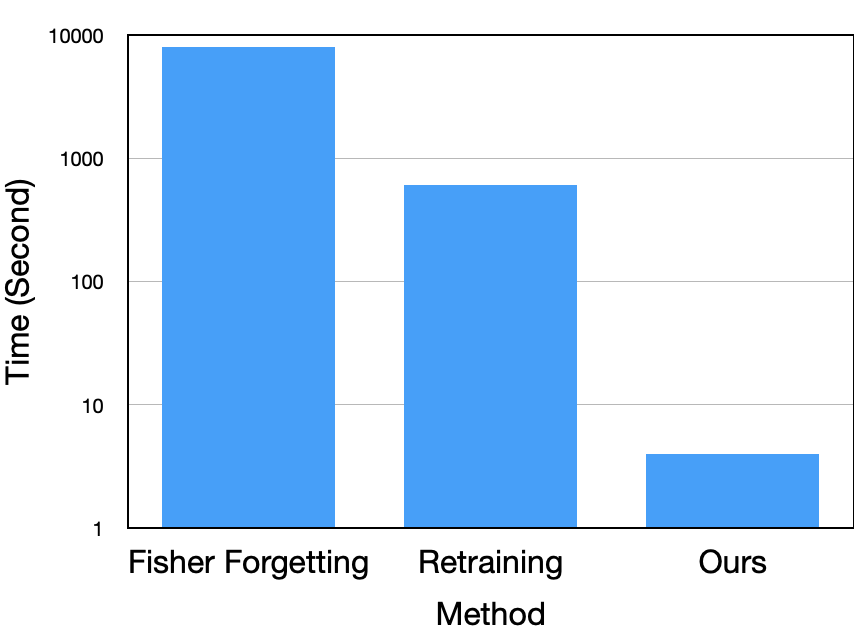}
    \caption{Figure shows the training time comparison between Retraining, Fisher Forgetting, and UNSIR (our method).}
    \label{fig:efficiency_analysis}
\end{figure}

Our method is fast and highly efficient in comparison to retraining and the existing unlearning approaches~\cite{golatkar2020eternal,golatkar2020forgetting}. The Fisher Forgetting~\cite{golatkar2020eternal} and NTK based forgetting~\cite{golatkar2020forgetting} approaches require Hessian approximation which is computationally very expensive. These methods give some bounds on the amount of information remaining but they are quite inefficient for practical use. They take even more time than retraining itself. Whereas, retraining took us around 10 minutes (617 seconds), it took us more than 2 hours to run Fisher forgetting~\cite{golatkar2020eternal} for 1-class unlearning in ResNet18+CIFAR-10. The Fisher forgetting for 1-class unlearning in AllCNN+CIFAR-10 takes around 1 hour. For CIFAR-100, the estimated time surpassed 25 hours. The NTK based forgetting~\cite{golatkar2020forgetting} uses Fisher noise along with NTK based model approximations and thus is even more time consuming. Our method only requires \textit{1.1 seconds} for 40 steps of noise optimization on ResNet18+CIFAR-10, \textit{1.70 seconds} for one epoch of impair, and \textit{1.13 seconds} for an epoch of repair. The total computational time for unlearning is \textit{less than 4 seconds}. This is 154$\times$ faster than the retraining approach, 1875$\times$ faster than the Fisher approach. We achieve fast unlearning without compromising the effectiveness of the method. Moreover,~\textit{our method is scalable} to large problems and big models. The cost of noise matrix estimation depends on the cost of a forward pass in the model. Usually, in multi-class unlearning, the cost of noise matrix estimation is linearly dependent on the number of forget classes. In case of UNSIR, the algorithm is executed only once for both single-class or multi-class unlearning. Thus, our method offers the most efficient multiple class unlearning among them. Fig. \ref{fig:efficiency_analysis} shows the time complexity comparison for retraining, Fisher Forgetting and UNSIR. Our method requires 1250$\times$ less time than retraining and 125$\times$ less time than Fisher forgetting.

\subsection{Comparing Different Impair-Repair Configurations}
We conduct experiments to provide a comparison between different impair-repair configurations on ResNet18+VGGFace-100 in Table~\ref{tab:impair-repair}. A single impair-repair cycle does not yield the expected $0\%$ accuracy on forget set. Since most of the damage is done in the impair step, we observe the effect of executing two impair steps before the repair step. After impair the performance on the forget set reaches the desired $0\%$ but the model regains $3\%$ accuracy after the repair step. We then execute two cycles of impair-repair. This means one impair, one repair, one impair, and one repair step. This yields the expected $0\%$ on the forget set with minimal loss on performance on the retain set ($72.79$,$72.5$ vs $70.86\%$). Furthermore, additional ablation analysis is presented in the supplementary material.

\begin{table}[]
\footnotesize
\centering
\caption{Observing the effect of different combination of impair-repair steps. The experiments are done on ResNet18+VGGFace-100}
\resizebox{\columnwidth}{!}{
\begin{tabular}{c|c|cc}
\hline
{\multirow{2}{*}{Setting}} & {\multirow{2}{*}{Intermediate}}& Accuracy on  & Accuracy On\\
{} & {} & Forget Set & Retain Set \\
\hline
\multirow{3}{*}{1$\times$(Impair-Repair)} & Before Impair& 80.63 & 94\\
\cline{2-4}
{} & After Impair & 2 & 53\\
\cline{2-4}
 & After Repair & 3 & 72.79 \\

\hline
\multirow{3}{*}{\shortstack{2$\times$(Impair)-1$\times$(Repair)}} & Before Impair& 80.63 & 94\\
\cline{2-4}
 & After 2 Impairs & 0 & 49.05\\
\cline{2-4}
 & After Repair & 3 & 72.5\\
\hline

\multirow{3}{*}{\shortstack{2$\times$(Impair-Repair)}} & {Before Impair} & 80.63 & 94\\
\cline{2-4}
{} & {After Cycle 1} & 3 & 72.79\\
\cline{2-4}
{} & {After Cycle 2} & 0 & 70.86\\
\hline
\end{tabular}
}
\label{tab:impair-repair}
\end{table}

\subsection{Limitations}
Our method achieves unlearning in already trained deep learning models. The existing approaches~\cite{golatkar2020eternal,guo2020certified,golatkar2020forgetting,golatkar2021mixed} either require training the models in a specific manner or make impractical assumptions like linear models or treating deep model training as a convex optimization problem and are thus incompatible with our target settings of unlearning from an already trained model. The unlearning approach in~\cite{bourtoule2021machine} provides exact unlearning guarantee but consumes lot of memory and requires implementation during the training process. Our method can be used to perform unlearning as an afterthought i.e., delete data from previously deployed deep learning models. Similarly, unlike the linear/convex case, where strong bounds on the amount of remaining information can be formulated, forgetting on DNNs often does not come with any provable bound. This is still an open problem. The kind of information bounds given in the above works are not compatible with our framework. To cope with this limitation, we conduct extensive experimental analysis to check the unlearning performance through a variety of widely accepted metrics. We use performance on retain and forget set, layer-wise weight difference, prediction distribution comparison for forget set and relearn time to evaluate the unlearning and showcase that our method is effective with no empirical signs of information leakage. However, a more formal guarantee of unlearning might be desired in highly privacy sensitive applications.\par
Unlearning a random cohort of data is beyond the scope of this work. Although, in theory an error-maximizing noise matrix can be generated corresponding to the random samples. But this would hurt the zero-glance assumption and thus, unlearning random samples, or only a subset of a class is out of the scope of this paper. Furthermore, the analysis of adaptive adversaries, that have exact knowledge about the proposed algorithm, is out of scope as well. We also point out the trade-off between the speed and accuracy in our unlearning method. For example, the efficiency gain in the proposed method is 154$\times$ and 1875$\times$ more than the retrain and Fisher method, respectively. However, this efficiency gain comes at the cost of decreased accuracy in comparison to the retrain method.

\section{Conclusion}
In this paper, we presented a stringent \textit{zero-glance} setting for unlearning and explore an efficient solution for it. We also develop a scalable, multiple class unlearning method. The unlearning method consists of learning an error-maximizing noise matrix followed by single pass impair and repair to update the network weights. Different from existing works, our method is highly efficient in unlearning multiple classes of data and we empirically demonstrate its effectiveness in a variety of deep networks such as CNN and Vision Transformer. The method is applicable to deep networks trained with any kind of optimization. Excellent unlearning results on a large-scale face recognition dataset is also shown which is a first such attempt. Our work opens up a new direction for efficient multi-class unlearning on large-scale problems. A possible future direction could be to perform unlearning without using any kind of training samples. 

\section*{Acknowledgment}
This research/project is supported by the National Research Foundation, Singapore under its Strategic Capability Research Centres Funding Initiative. Any opinions, findings and conclusions or recommendations expressed in this material are those of the author(s) and do not reflect the views of National Research Foundation, Singapore.

{
    \bibliographystyle{IEEEtran}
    \bibliography{main}
}

\newpage
\onecolumn
\appendix

\section{Additional Ablation Analysis}
\subsection{Effect of using Different Learning Rates in UNSIR}
The learning rate in the impair and repair steps of UNSIR plays a crucial role in determining the unlearning outcomes. The impair step is likely to damage the feature mapping in the model corresponding to the forget class. The repair step recovers the performance on the retain classes. Tables~\ref{impair lr} shows the result for varying learning rate in impair step while keeping the learning rate fixed in the repair step (at 0.01). Similarly, Table~\ref{repair lr} shows the result for varying learning rate in repair step while keeping the learning rate fixed in the impair step (at 0.02). We observe that the learning rate 0.01 and 0.001 offer a reasonable trade-off between the performance in the forget and retain classes. Similar to the general machine learning methods, choosing an optimal learning rate in UNSIR is part of the hyper parameter tuning process in machine unlearning.

\begin{table*}[!htb]
\footnotesize
\centering
\caption{Single class unlearning at various learning rates in the \textit{impair step}. The learning rate for \textit{repair step} is fixed at \textbf{0.01}}
\begin{tabular}{c|ccccc}
\hline
Model & Learning Rate (Impair) & Metrics& Original Model & After Impair & After Repair\\
\hline
\multirow{8}{*}{ResNet18} & \multirow{2}{*}{1} & $A_{D_r}$ $\uparrow$ & 77.86 & 29.38 & 37.86\\
{} & {} & $A_{D_f}$ $\downarrow$ & 81.01 & 0 & 0\\
\cline{2-6}

 & \multirow{2}{*}{0.1} & $A_{D_r}$ $\uparrow$ & 77.86 & 24.17 & 41.07\\
{} & {} & $A_{D_f}$ $\downarrow$ & 81.01 & 1.70 & 0\\
\cline{2-6}
 & \multirow{2}{*}{0.01} & $A_{D_r}$ $\uparrow$ & 77.86 & 68.28 & 71.47\\
{} & {} & $A_{D_f}$ $\downarrow$ & 81.01 & 1.1 & 0\\
\cline{2-6}
 & \multirow{2}{*}{0.001} & $A_{D_r}$ $\uparrow$ & 77.86 & 77.75 & 69.45\\
{} & {} & $A_{D_f}$ $\downarrow$ & 81.01 & 72.06 & 0.80\\
\hline
\multirow{8}{*}{AllCNN} & \multirow{2}{*}{1} & $A_{D_r}$ $\uparrow$ & 82.64 & 10.51 & 14.91\\
{} & {} & $A_{D_f}$ $\downarrow$ & 91.02 & 0& 0\\
\cline{2-6}
 & \multirow{2}{*}{0.1} & $A_{D_r}$ $\uparrow$ & 82.64 & 14.47 & 28.5 \\
{} & {} & $A_{D_f}$ $\downarrow$ & 91.02 & 0& 0\\
\cline{2-6}
 & \multirow{2}{*}{0.01} & $A_{D_r}$ $\uparrow$ & 82.64 & 75.50& 70.38\\
{} & {} & $A_{D_f}$ $\downarrow$ & 91.02 & 2.41& 0\\
\cline{2-6}
 & \multirow{2}{*}{0.001} & $A_{D_r}$ $\uparrow$ & 82.64 & 85.42& 74.61\\
{} & {} & $A_{D_f}$ $\downarrow$ & 91.02 & 77.00& 3.35\\
\hline
\end{tabular}
\label{impair lr}
\end{table*}

\begin{table}[!htb]
\footnotesize
\centering
\caption{Single class unlearning at various learning rates in the \textit{repair step}. The learning rate for \textit{impair step} is fixed at \textbf{0.02}}
\begin{tabular}{c|ccccc}
\hline
Model & Learning Rate (Repair)& Metrics& Original Model & After Impair & After Repair \\
\hline
\multirow{8}{*}{ResNet18}& \multirow{2}{*}{1} & $A_{D_r}$ $\uparrow$ & 77.86 & 63.75 & 11.03\\
{} & {} & $A_{D_f}$ $\downarrow$ & 81.01 & 0.5 & 0\\
\cline{2-6}
 & \multirow{2}{*}{0.1} & $A_{D_r}$ $\uparrow$ & 77.86 & 63.75 & 30.85\\
{} & {} & $A_{D_f}$ $\downarrow$ & 81.01 & 0.5 & 0\\
\cline{2-6}
 & \multirow{2}{*}{0.01} & $A_{D_r}$ $\uparrow$ & 77.86 & 63.75 & 71.31\\
{} & {} & $A_{D_f}$ $\downarrow$ & 81.01 & 0.5 & 0\\
\cline{2-6}
 & \multirow{2}{*}{0.001} & $A_{D_r}$ $\uparrow$ & 77.86 & 63.75 & 71.18\\
{} & {} & $A_{D_f}$ $\downarrow$ & 81.01 & 0.5 & 0.1\\
\hline
\multirow{8}{*}{AllCNN} & \multirow{2}{*}{1} & $A_{D_r}$ $\uparrow$ & 82.64 & 67.62 & 10.19\\
{} & {} & $A_{D_f}$ $\downarrow$ & 91.02 & 0.29 & 0\\
\cline{2-6}
 & \multirow{2}{*}{0.1} & $A_{D_r}$ $\uparrow$ & 82.64 & 67.62 & 13.94 \\
{} & {} & $A_{D_f}$ $\downarrow$ & 91.02 & 0.29 & 0 \\
\cline{2-6}
 & \multirow{2}{*}{0.01} & $A_{D_r}$ $\uparrow$ & 82.64 & 67.62 & 73.62 \\
{} & {} & $A_{D_f}$ $\downarrow$ & 91.02 & 0.29 & 0\\
\cline{2-6}
 & \multirow{2}{*}{0.001} & $A_{D_r}$ $\uparrow$ & 82.64 & 67.62 & 80.9\\
{} & {} & $A_{D_f}$ $\downarrow$ & 91.02 & 0.29 & 0.3\\
\hline
\end{tabular}
\label{repair lr}
\end{table}

\subsection{Different Levels of Weight Penalization}
The $\lambda$ value controls the amount of penalization imposed on the weights while learning the noise matrix. The change in this value governs the behaviour of the final noise matrix. We obtain the unlearning results for different values of $\lambda$ and present the same in Table~\ref{lambda}. In case of no penalization, i.e. $\lambda=0$, we obtain good unlearning performance (acc on $D_{f}=0$) but get sub-optimal performance over the retain set (acc on $D_{r}=67.03$) for ResNet18. Similarly, for AllCNN, there is a difference between the performance with and without penalization, although the difference is not as significant as in ResNet18. In general, penalization with some $\lambda$ value gives better results in both AllCNN and ResNet18 in terms of expected performance over the forget and retain set. 

\subsection{Using Different Proportions of Retain Data ($D_{r}$) for Unlearning}
Our unlearning method requires only small amount of training data ($\sim10\%-20\%$) from the retain set $D_{r}$. We explore the effect of using higher proportions of samples on the overall performance of the model. In Table~\ref{percentage-train}, we report the performance obtained with 1\%, 10\%, 20\%, 30\%, 40\%, 50\%, 60\% proportions of retain set. The results are in line with the common understanding that with the availability of more data, we can achieve slightly better unlearning results.

\begin{table}[]
\centering
\caption{The effect of different values of $\lambda$ parameter in L2 weight penalization. The results are shown for 1-class unlearning.}
\begin{tabular}{c|c|cccc}
\hline
Model & $\lambda$ & Metrics & Original Model & After Unlearning & Relearn Time (RT)\\
\hline
\multirow{12}{*}{ResNet18} & \multirow{2}{*}{10} & $A_{D_r}$ $\uparrow$ & 77.86 & 70.96 & \multirow{2}{*}{$>$100}\\
{} & {} & $A_{D_f}$ $\downarrow$ & 81.01& 0 & {}\\
\cline{2-6}
& \multirow{2}{*}{1} & $A_{D_r}$ $\uparrow$ & 77.86 & 68.34 & \multirow{2}{*}{92}\\
{} & {} & $A_{D_f}$ $\downarrow$ & 81.01 & 0 & {}\\
\cline{2-6}
 & \multirow{2}{*}{0.1} & $A_{D_r}$ $\uparrow$& 77.86 & 71.06 & \multirow{2}{*}{90}\\
{} & {} & $A_{D_f}$ $\downarrow$ & 81.01 & 0 & {}\\
\cline{2-6}
 & \multirow{2}{*}{0.01} & $A_{D_r}$ $\uparrow$& 77.86 & 70.04 & \multirow{2}{*}{$>$100}\\
{} & {} & $A_{D_f}$ $\downarrow$ & 81.01 & 0 & {}\\
\cline{2-6}
 & \multirow{2}{*}{0.001} & $A_{D_r}$  $\uparrow$ & 77.86 & 71.47 & \multirow{2}{*}{$>$100}\\
{} & {} & $A_{D_f}$ $\downarrow$ & 81.01 & 0 & {}\\
\cline{2-6}
 & \multirow{2}{*}{0} & $A_{D_r}$  $\uparrow$ & 77.86 & 67.03 & \multirow{2}{*}{$>$100}\\
{} & {} & $A_{D_f}$ $\downarrow$ & 81.01 & 0 & {}\\
\hline
\multirow{12}{*}{AllCNN} & \multirow{2}{*}{10} & $A_{D_r}$ $\uparrow$ & 82.64 & 78.21 & \multirow{2}{*}{$>$100}\\
{} & {} & $A_{D_f}$ $\downarrow$ & 91.02 & 0 & {}\\
\cline{2-6}
 & \multirow{2}{*}{1} & $A_{D_r}$ $\uparrow$ & 82.64 & 76.92 & \multirow{2}{*}{$>$100}\\
{} & {} & $A_{D_f}$ $\downarrow$ & 91.02 & 0 & {}\\
\cline{2-6}
 & \multirow{2}{*}{0.1} & $A_{D_r}$ $\uparrow$& 82.64 & 76.16 & \multirow{2}{*}{$>$100}\\
{} & {} & $A_{D_f}$ $\downarrow$ & 91.02 & 0 & {}\\
\cline{2-6}
 & \multirow{2}{*}{0.01} & $A_{D_r}$ $\uparrow$& 82.64 & 74.28 & \multirow{2}{*}{$>$100}\\
{} & {} & $A_{D_f}$ $\downarrow$ & 91.02 & 0 & {}\\
\cline{2-6}
 & \multirow{2}{*}{0.001} & $A_{D_r}$ $\uparrow$ & 82.64 & 76.55 & \multirow{2}{*}{$>$100}\\
{} & {} & $A_{D_f}$ $\downarrow$ & 91.02 & 0 & {}\\
\cline{2-6}
& \multirow{2}{*}{0} & $A_{D_r}$ $\uparrow$ & 82.64 & 76.07 & \multirow{2}{*}{$>$100}\\
{} & {} & $A_{D_f}$ $\downarrow$ & 91.02 & 0 & {}\\
\hline
\end{tabular}
\label{lambda}
\end{table}

\begin{table}[]
\centering
\caption{Single class unlearning in ResNet18 with different proportion of retain samples used from the retain set $D_{r}$.}
\begin{tabular}{ccccc}
\hline
\# Samples(\% of $D_{r}$)  & Metrics & Original Model& After Impair & After Repair\\
\hline
\multirow{2}{*}{50 (1\%)} & $A_{D_r}$ $\uparrow$ & 77.86 & 13.95 & 34.81\\
{} & $A_{D_f}$ $\downarrow$ & 81.01 & 43.43 & 0.89\\
\hline
\multirow{2}{*}{500 (10\%)} & $A_{D_r}$ $\uparrow$ & 77.86 & 51.05 & 64.45\\
{} & $A_{D_f}$ $\downarrow$ & 81.01 & 0.29 & 0\\
\hline
\multirow{2}{*}{1000 (20\%)} & $A_{D_r}$ $\uparrow$ & 77.86 & 66.22 & 67.05\\
{} & $A_{D_f}$ $\downarrow$ & 81.01 & 0.69 & 0 \\
\hline
\multirow{2}{*}{1500 (30\%)} & $A_{D_r}$ $\uparrow$ & 77.86 & 68.26 & 70.82\\
{} & $A_{D_f}$ $\downarrow$ & 81.01 & 0.78 & 0\\
\hline
\multirow{2}{*}{2000 (40\%)} & $A_{D_r}$ $\uparrow$ & 77.86 & 68.43 & 71.91\\
{} & $A_{D_f}$ $\downarrow$ & 81.01 & 0 & 0\\
\hline
\multirow{2}{*}{2500 (50\%)} & $A_{D_r}$ $\uparrow$ & 77.86 & 70.90 & 73.07\\
{} & $A_{D_f}$ $\downarrow$ & 81.01 & 0 & 0\\
\hline
\multirow{2}{*}{3000 (60\%)} & $A_{D_r}$ $\uparrow$ & 77.86 & 72.25 & 74.03\\
{} & $A_{D_f}$ $\downarrow$ & 81.01 & 0.09 & 0\\
\hline
\end{tabular}
\label{percentage-train}
\end{table}

\subsection{Sequential Forgetting}
\begin{figure}[]
 \centering
 \includegraphics[width=0.6\linewidth]{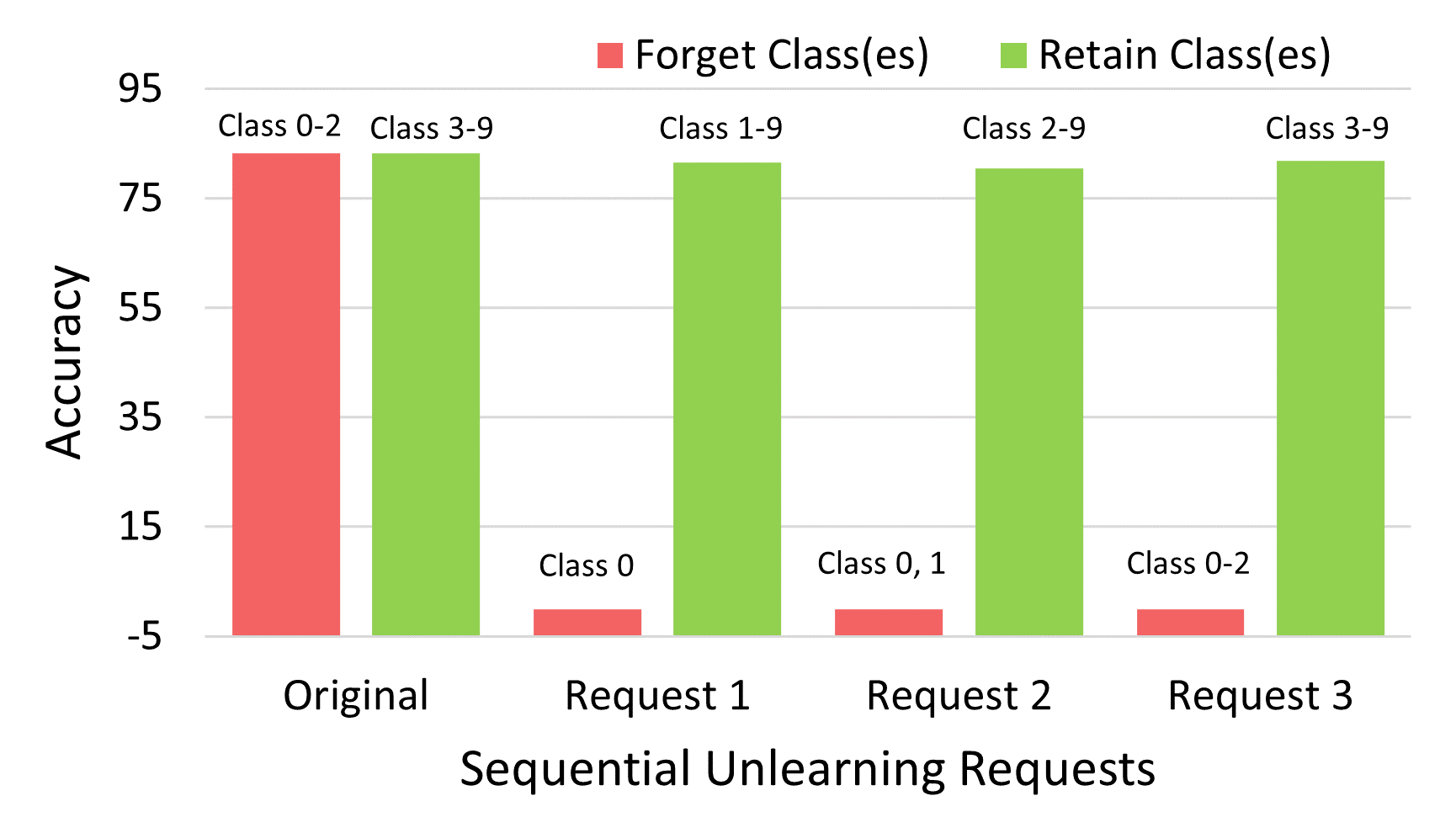}
  \caption{Performance of our method in sequential unlearning requests. We sequentially forget class-0, class-1, and class-2, respectively.}
  \label{fig:sequential_requests}
\end{figure}

In practical scenario, several sequential requests to unlearn may be raised. We show a sample case of 3 sequential unlearning requests in CIFAR-10 for AllCNN model in Fig.~\ref{fig:sequential_requests}.
We observe that the performance of the model does not deteriorate much after several sequential requests.

\subsection{Healing after Multiple Steps of Repair}
\begin{figure}[]
    \centering
    \includegraphics[width=.4\linewidth]{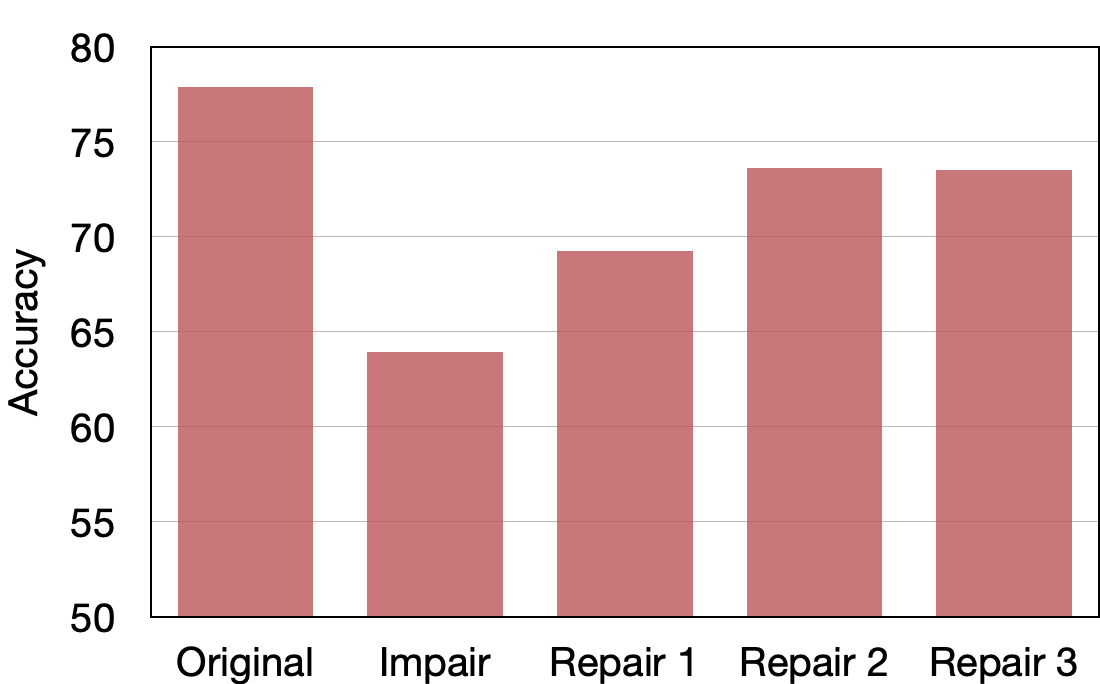}
    \caption{Figure shows the effect of running multiple repair steps for 1-class unlearning in ResNet18.}
    \label{fig:healing_analysis}
\end{figure}

Beyond single step repair, we explore to check if further performance improvement is possible with multiple steps of repair. The accuracy gain on $D_r$ after 1, 2 and 3 steps of repair (ResNet18 on CIFAR-10) is given in Fig.~\ref{fig:healing_analysis}. Although most of the performance gain can be obtained by a single step of repair, however multiple steps can be leveraged to improve the performance further on $D_r$. The number of repair steps act as a trade-off between the computation cost and the performance on $D_r$.

\section{On the Validity of Retrained Model as the Gold Standard} 
In the literature~\cite{golatkar2020eternal,golatkar2020forgetting,golatkar2021mixed}, the model retrained only on the retain set of data ($D_{r}$) is treated as the \textit{gold standard model}. The existing methods~\cite{golatkar2020eternal,golatkar2020forgetting,golatkar2021mixed} are designed to move the network weights closer to such a 'gold standard'. However, the parameters in a deep network can have many degrees of freedom~\cite{gao2016degrees}. The requested information can be erased by perturbing the parameters in any direction. Thus, there may exist multiple network parameter configurations that can achieve effective unlearning. The retrained model is one among such possible set of models. Therefore, it may not be appropriate to denote the retrained model as the only 'gold standard'. In Table~\ref{dist} we report the Euclidean distance between the respective layers of the proposed unlearned model and the retrained (gold standard) model. We observe that even though our method performs well on all the readout functions, it is quite far from the retrain model in the parameter space. This suggests that the unlearning can be achieved even without moving the network weights toward the gold standard model. \textit{Thus, proximity to the retrained model may reveal the unlearned model but models lying far away from it can be equally good candidates.} We posit there exist multiple gold standard models for unlearning. The proposed method suggests the existence of such models.

\begin{table*}[!htb]
\centering
\caption{We measure the Euclidean distances between parameter set of the proposed \textit{unlearned model}, the \textit{original model}, and the \textit{retrain model} on CIFAR-10}
\resizebox{\columnwidth}{!}{
\begin{tabular}{c|c|c|c|c|c|c}
\hline
{} & \multicolumn{3}{|c|}{Unlearning in 1 class} & \multicolumn{3}{c}{Unlearning in 7 classes}\\
\hline
Model & \textit{dist(original,proposed)} & \textit{dist(original,retrain)} & \textit{dist(proposed,retrain)} & \textit{dist(original,proposed)} & \textit{dist(original,retrain)} & \textit{dist(proposed,retrain)}\\
\hline
ResNet18 & 164.06 & 81.96 & 185.14 & 91.53 & 70.59 & 116.93\\
\hline
AllCNN & 41.44 & 78.38 & 92.5 & 34.85 & 69.39 & 78.29\\
\hline
\end{tabular}
}
\label{dist}
\end{table*}



\end{document}